\documentclass[10pt,twocolumn,letterpaper]{article}

\usepackage{cvpr}              

\usepackage{subcaption}
\usepackage{float}
\usepackage[margin=1in]{geometry}  
\usepackage{caption}
\usepackage{multirow}

\usepackage{nicefrac}       
\usepackage{microtype}      

\usepackage{graphicx}
\usepackage{caption}

\usepackage{amsmath}
\usepackage{amsfonts}
\usepackage{amssymb}
\usepackage{mathtools}
\usepackage{amsthm}
\usepackage{multirow}
\usepackage{dsfont}
\usepackage{algorithm}
\usepackage{algpseudocode}
\usepackage[T1]{fontenc}

\usepackage{bbold}
\usepackage{thmtools} 
\usepackage{thm-restate}
\usepackage{bm}

\theoremstyle{plain}
\newtheorem{theorem}{Theorem}[section]

\theoremstyle{definition}
\newtheorem{definition}[theorem]{Definition}

\theoremstyle{remark}

\usepackage{xcolor}

\usepackage{pdfpages}

\setcitestyle{square,numbers,comma}

\usepackage[inline]{enumitem}
\usepackage{makecell}
\usepackage{multirow}
\usepackage[export]{adjustbox}

\newcommand{\ourmethod}{\textsc{DiTASK}}

\usepackage{enumitem}%

\newcommand{\stepsfont}{\normalfont}%

\usepackage{enumitem}
\newlist{steps}{enumerate}{9}
\setlist[steps]{before=\upshape\stepsfont,font=\ttfamily} %
\setlist[steps,1]{leftmargin=*,labelsep=1ex,label={\arabic*.},ref={\arabic*}} %
\setlist[steps,2]{leftmargin=*,labelsep=1ex,label={\Alph*.},ref={\thestepsi.\Alph*}} %
\setlist[steps,3]{leftmargin=*,labelsep=1ex,label={\arabic*.},
  ref={\thestepsii.\arabic*}} %
\setlist[steps,4]{leftmargin=*,labelsep=1ex,label={\alph*.},
  ref={\thestepsiii.\alph*}} %
\setlist[steps,5]{leftmargin=*,labelsep=1ex,label={\arabic*.},
  ref={\thestepsiv.\arabic*}} %
\setlist[steps,6]{leftmargin=*,labelsep=1ex,label={\Alph*.},
  ref={\thestepsv.\Alph*}} %
\setlist[steps,7]{leftmargin=*,labelsep=1ex,label=\arabic*.,
  ref={\thestepsvi.\roman*}} %
\setlist[steps,8]{leftmargin=*,labelsep=1ex,label={\alph*.},
  ref={\thestepsvii.\alph*}} %
\setlist[steps,9]{leftmargin=*,labelsep=1ex,label={\arabic*.},
  ref={\thestepsviii.\arabic*}} %
\newlist{stepitems}{itemize}{1} 
\setlist[stepitems]{before=\upshape\ttfamily,label=*,leftmargin=*,labelsep=1ex}%

\makeatletter
\define@key{Gin}{align}[]{%
}
\makeatother

\usepackage{amsmath,amsfonts,bm}

\def\1{\bm{1}}

\def\rvh{{\mathbf{h}}}
\def\rvu{{\mathbf{i}}}

\def\rvu{{\mathbf{u}}}
\def\rvv{{\mathbf{v}}}

\def\rvx{{\mathbf{x}}}

\def\rmA{{\mathbf{A}}}
\def\rmB{{\mathbf{B}}}

\def\rmU{{\mathbf{U}}}
\def\rmV{{\mathbf{V}}}
\def\rmW{{\mathbf{W}}}

\def\vtheta{{\bm{\theta}}}

\DeclareMathAlphabet{\mathsfit}{\encodingdefault}{\sfdefault}{m}{sl}
\SetMathAlphabet{\mathsfit}{bold}{\encodingdefault}{\sfdefault}{bx}{n}

\def\gM{{\mathcal{M}}}
\def\gN{{\mathcal{N}}}
\def\gO{{\mathcal{O}}}
\def\gP{{\mathcal{P}}}

\def\sR{{\mathbb{R}}}

\definecolor{cvprblue}{rgb}{0.21,0.49,0.74}
\usepackage[pagebackref,breaklinks,colorlinks,allcolors=cvprblue]{hyperref}
\usepackage[capitalize,noabbrev]{cleveref}

\newcommand{\revision}[1]{{\color{black}#1}}
\def\paperID{4247} 
\def\confName{CVPR}
\def\confYear{2025}

\title{\textsc{DiTASK}: Multi-Task Fine-Tuning with Diffeomorphic Transformations}

\author{Krishna Sri Ipsit Mantri\\
Purdue University\\
West Lafayette, USA\\
{\tt\small mantrik@purdue.edu}
\and
Carola-Bibiane Sch\"onlieb\\
University of Cambridge\\
Cambridge, UK\\
{\tt\small cbs31@cam.ac.uk}
\and
Bruno Ribeiro\\
Purdue University\\
West Lafayette, USA\\
{\tt\small ribeirob@purdue.edu}
\and
Chaim Baskin\\
Ben-Gurion University of the Negev\\
Beer Sheva, Israel\\
{\tt\small chaimbaskin@bgu.ac.il}
\and 
Moshe Eliasof\\
University of Cambridge\\
Cambridge, UK\\
{\tt\small me532@cam.ac.uk}
}
\usepackage{microtype}
\usepackage{times}
\begin{document}

\maketitle
\begin{abstract}
Pre-trained Vision Transformers now serve as powerful tools for computer vision. Yet, efficiently adapting them for multiple tasks remains a challenge that arises from the need to modify the rich hidden representations encoded by the learned weight matrices, without inducing interference between tasks. Current parameter-efficient methods like LoRA, which apply low-rank updates, force tasks to compete within constrained subspaces, ultimately degrading performance. We introduce \ourmethod\, a novel Diffeomorphic Multi-Task Fine-Tuning approach that maintains pre-trained representations by preserving weight matrix singular vectors, while enabling task-specific adaptations through neural diffeomorphic transformations of the singular values. By following this approach, \ourmethod\ enables both shared and task-specific feature modulations with minimal added parameters.  Our theoretical analysis shows that \ourmethod{} achieves full-rank updates during optimization, preserving the geometric structure of pre-trained features, and establishing a new paradigm for efficient multi-task learning (MTL). Our experiments on PASCAL MTL and NYUD show that \ourmethod\ achieves state-of-the-art performance across four dense prediction tasks, using \(75\%\) fewer parameters than existing methods. Our code is available  \href{https://github.com/ipsitmantri/DiTASK}{here}.
\end{abstract}    
\section{Introduction}
\label{sec:intro}

\begin{figure}[t]
    \centering
\includegraphics[width=0.9\linewidth]{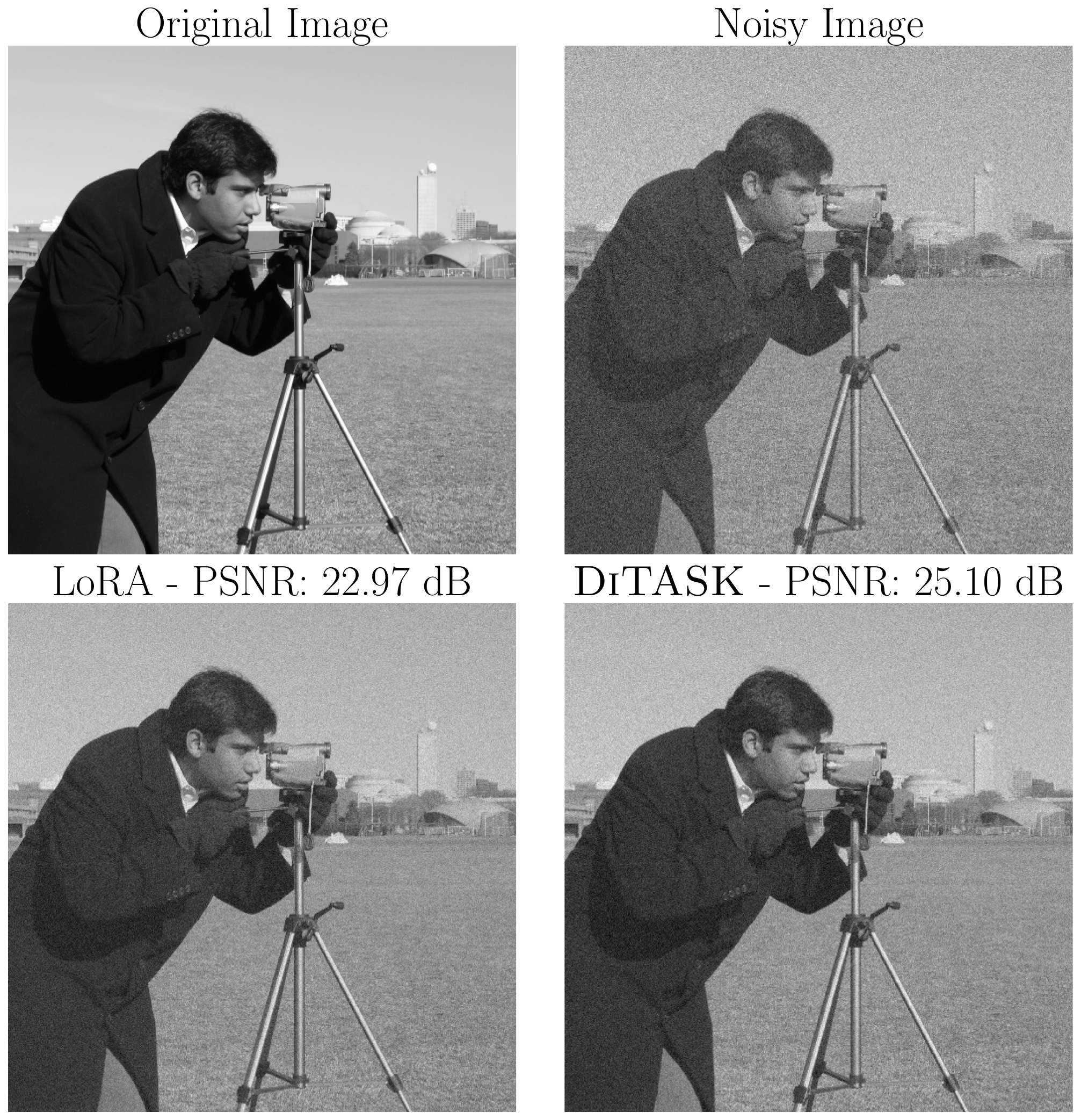}
    \caption{Image recovery comparison of LoRA and \ourmethod. Preserving singular vectors enables \ourmethod{} to achieve a higher peak signal-to-noise ratio (PSNR).}

    \label{fig:motivation}
\end{figure}

Vision Transformers (ViTs) have emerged as compelling models for computer vision, achieving remarkable performance across various tasks by leveraging self-attention mechanisms~\citep{vaswani2017attention,dosovitskiy2021an} and large-scale pre-training~\citep{steiner2021train, russakovsky2015imagenet} on extensive datasets like ImageNet-21k~\citep{deng2009imagenet}. These models encode essential visual knowledge in their weight matrices through singular vectors that characterize the principal directions in the input space and output spaces~\citep{NEURIPS2018_a7a3d70c, saxe2014exactsolutionsnonlineardynamics}.
\begin{figure}[!ht]
    \centering
\includegraphics[width=
\linewidth]{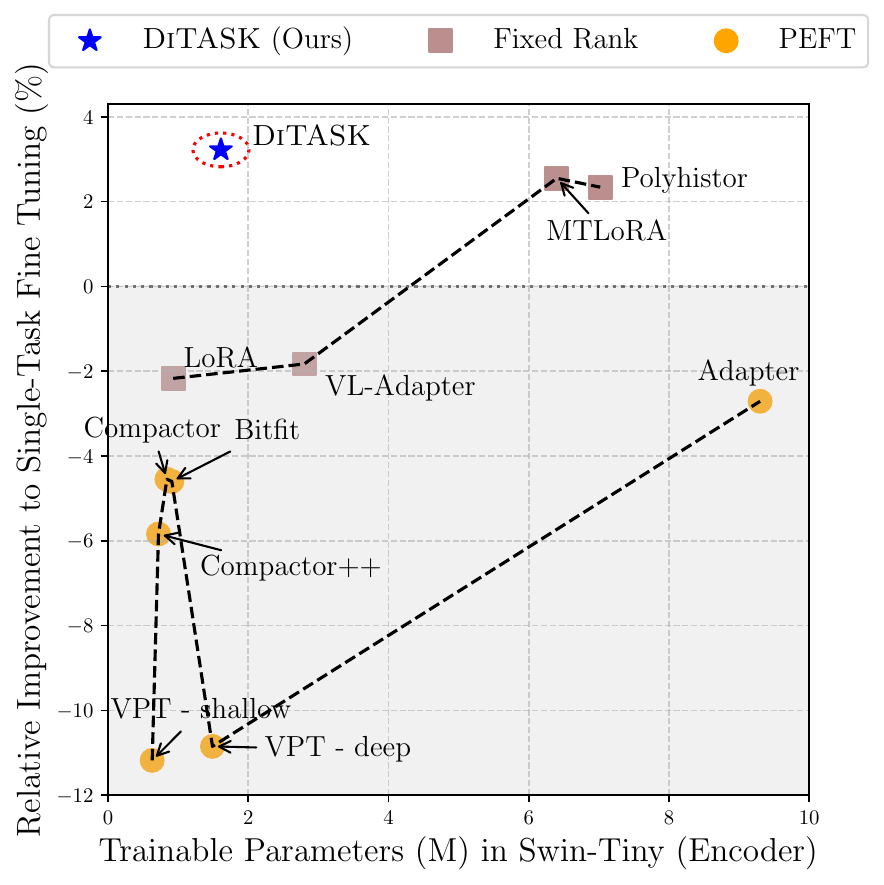}
    \caption{Performance–efficiency trade-off in multi-task learning (MTL) on PASCAL MTL tasks. \ourmethod\ outperforms single-task fine-tuning with fewer parameters than fixed-rank methods (LoRA~\citep{hu2022lora}) and other PEFT approaches (Adapter~\citep{houlsby2019parameterefficienttransferlearningnlp}). The gray region shows negative relative improvement over single-task tuning, highlighting the challenge of achieving gains under tight parameter budgets.}

    \label{fig:teaser}
\end{figure}
However, existing fine-tuning methods based on low-rank adaptation (LoRA) \cite{hu2022lora, agiza2024mtlora} may not preserve these learned singular vectors. In \Cref{fig:motivation}, we illustrate the importance of preserving the singular vectors; recovering a noisy image by adjusting only its singular values while keeping singular vectors yields \emph{superior} performance than using LoRA, which may not preserve the singular vectors of the image.  \revision{Singular vectors encode crucial image details, and low-rank denoising methods can lose these details, resulting in poor denoising quality. Preserving singular vectors serves as an implicit regularizer}. This motivating example suggests that useful adaptations can be achieved through singular value modulation and inspires us to design a mathematically-ground method to achieve this goal.

Fine-tuning of pre-trained ViTs for downstream tasks, such as few-shot learning~\citep{Hu_2022_CVPR,Park_2024_CVPR}, typically involves adapting the learned representations to the target application. However, the substantial size of these models calls for a resource-efficient fine-tuning technique. Parameter-efficient and low-rank optimization approaches, such as LoRA~\citep{hu2022lora}, offer promising solutions by reducing the number of trainable parameters. This strategy not only enhances resource efficiency (e.g., lower VRAM and FLOPS requirements), 
but also helps keep valuable pre-trained filters that capture essential visual knowledge~\citep{burns2023makespretrainedvisualrepresentations} by preventing catastrophic forgetting~\citep{luo2023investigatingforgettingpretrainedrepresentations, he2023preservingpretrainedfeatureshelps}.

In the context of single-task fine-tuning, parameter-efficient and low-rank methods have proven highly successful~\citep{hu2022lora, houlsby2019parameterefficienttransferlearningnlp, mahabadi2021parameter, karimi2021compacter}.
However, in {\textit{\textbf{Multi-Task Learning (MTL)}}} settings, these methods were shown to struggle \cite{ agiza2024mtlora}. Particularly, MTL typically requires isolating task gradients to prevent task interference (e.g., enforcing gradient orthogonality~\citep{pcgrad,NIPS1995_bdb106a0}), a requirement that becomes increasingly challenging as the dimension or rank of the gradients decreases.
Thus, we face a trade-off between lowering the number of trainable parameters and maintaining task performance. Evidently, as shown in  
\Cref{fig:teaser}, existing methods can offer performance enhancement in MTL settings only when significantly increasing the number of trainable, effectively undermining their purpose of lightweight adaptation.

We conjecture, inspired by recent studies~\citep{NEURIPS2023_4d69c1c0} that identified an \textit{incremental learning phenomenon} in transformers, that the underperformance of LoRA methods in MTL settings happens because of the direct and non-principled modification of weight matrices' singular vectors. These updates force distinct tasks to compete within the same constrained subspace. The key insight of our approach is realizing that while low-rank representations offer sufficient dimensionality for individual tasks, constraining multiple tasks to share the same low-rank subspace creates interference. An optimal subspace for one task may be suboptimal for others. Yet, existing methods force this competition by maintaining a single shared low-rank space, while a unique subspace for each task, like in MTLoRA~\citep{agiza2024mtlora}, may not take advantage of task synergies.

\noindent\textbf{Our approach.} 
To address the limitations of existing low-rank fine-tuning methods, we introduce \textbf{Diffeomorphic Multi-Task Adaptation} (\textbf{\ourmethod}), a novel approach leveraging neural diffeomorphisms parameterized by \textit{Continuous Piecewise Affine-Based} (CPAB) transformations~\citep{freifeld2015highly, freifeld2017transformations}. Specifically, \ourmethod\ preserves pre-trained feature patterns by retaining the singular vectors of learned weight matrices while dynamically adjusting singular values through neural diffeomorphic transformations at each optimization step. 
Our diffeomorphic transformations are parameterized by Continuous Piecewise Affine (CPA)~\cite{freifeld2015highly,freifeld2017transformations} velocity fields over the weight matrix domain, ensuring continuous, invertible deformations. This design enables adaptation across multiple tasks while preserving the knowledge embedded during pre-training.
We validate \ourmethod\ on the PASCAL MTL dataset~\citep{pascal}, a widely used benchmark for dense prediction tasks such as semantic segmentation, edge detection, and human body part segmentation. Our method achieves a \(26.27\%\) improvement in average task performance while reducing parameter usage by \(4 \times\) compared to state-of-the-art methods like MTLoRA~\citep{agiza2024mtlora}, demonstrating the effectiveness of preserving pre-trained feature directions for efficient multi-task fine-tuning.

\begin{figure*}[!t]
    \centering
    \begin{subfigure}[c]{0.6\linewidth}
        \centering
        \includegraphics[width=1.0\linewidth]{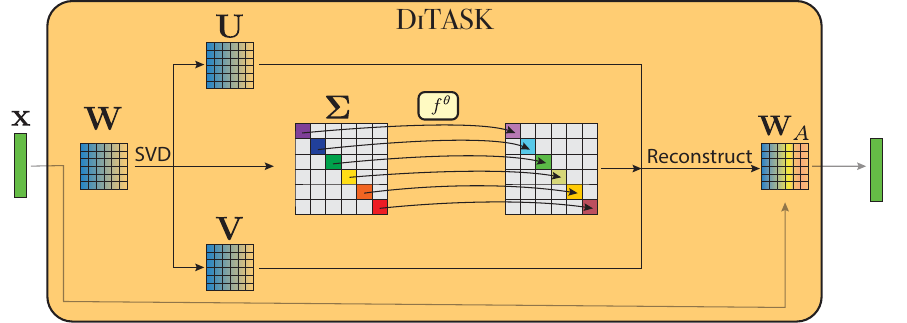}
        \label{fig:left}
    \end{subfigure}
    \hfill
    \begin{subfigure}[c]{0.3\linewidth}
        \centering
        \vspace{-0.75em}
        \includegraphics[width=0.8\linewidth]{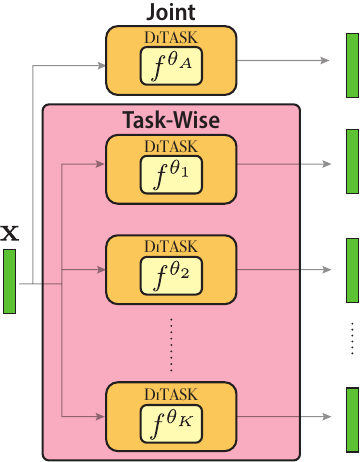}
        \label{fig:overview_right}
    \end{subfigure}
    \caption{\textbf{Overview of our \ourmethod\ within Multi-Task Learning (MTL).} (Left) Input features $\rvx$ are transformed via a modulated weight matrix $\rmW_A$, constructed by applying a neural diffeomorphism $f^{\vtheta}$ to the singular values of a pre-trained weight matrix $\rmW$, resulting in updated features $\rmW_A \rvx$. (Right) We use two sets of \ourmethod\ modules: a joint transformation module for learning task synergies, and task-specific modules for individual adaptations. The transformed outputs serve as inputs to a shared decoder. This setup enables flexible, parameter-efficient fine-tuning while preserving the singular vectors of pre-trained weights.}

    \label{fig:side_by_side}
\end{figure*}

\noindent
\textbf{Our contributions} are as follows:
\begin{itemize}
    \item We propose a novel approach to MTL fine-tuning that leverages neural diffeomorphisms for singular value adaptation, enabling fine-tuning weight updates while preserving pre-trained representation structure.
    
    \item We develop an efficient parametrization using learnable CPA velocity fields that require only 32 additional trainable parameters per layer.

    \item We theoretically analyze the properties of \ourmethod{}, showing that it requires less memory than other low-rank Transformer adaptation methods.
    
    \item \ourmethod{} demonstrates state-of-the-art performance on PASCAL MTL tasks, achieving \(26.27\%\) improvement over existing methods like MTLoRA. 
\end{itemize}
The rest of the paper is organized as follows: In \Cref{sec:relatedwork}, we discuss the related work on parameter-efficient fine-tuning and MTL. In \Cref{sec:background}, we provide the mathematical background to understand \ourmethod{}. \Cref{sec:method} details \ourmethod\, for learning diffeomorphic  singular value adjustment, and \Cref{sec:exp} presents experimental results and analyses. 
\section{Related Work}
\label{sec:relatedwork}
Our work lies at the intersection of four research areas: multi-task vision learning, parameter-efficient fine-tuning, neural network adaptation, and diffeomorphisms. We discuss each area with a focus on how they relate to weight matrix transformations and singular value adaptation. 
\\
\noindent\textbf{Multi-Task Learning in Vision.} 
Multi-task learning models in vision typically adopt an encoder-decoder structure, where a shared encoder captures common features, and task-specific decoders handle individual objectives~\citep{misra2016cross}. This approach has proven particularly effective for dense prediction tasks, with different decoders handling semantic segmentation, depth estimation, and surface normal prediction while sharing a common feature backbone~\citep{kendall2018multi}. While recent work has extended this framework to Vision Transformers (ViTs)~\citep{dosovitskiy2021an}, these adaptations face unique challenges due to their large parameter count and the complex interactions between task-specific updates. 
To address these issues, \ourmethod\ introduces an encoder adaptation strategy that modulates singular values while retaining pre-trained singular vectors, rather than modifying full weight matrices or adding task-specific layers. This leverages the encoder-decoder framework for efficient adaptation of the shared encoder.
\\
\noindent\textbf{Parameter-Efficient Fine-tuning Methods.} 
These methods reduce computational overhead for model adaptation using various strategies. Adapter-based approaches~\citep{houlsby2019parameterefficienttransferlearningnlp} insert small trainable modules between transformer layers. Visual Prompt Tuning (VPT)~\citep{jia2022visual} introduces trainable parameters in ViT input tokens, distinguishing it from standard adaptation approaches. Compactor~\citep{mahabadi2021parameter} and BitFit~\citep{zaken2021bitfit} add minimal trainable parameters, enabling lightweight gradients via Kronecker products of rank-one and low-rank matrices and fine-tuning only biases, respectively. Hyperformers~\citep{mahabadi2021parameter} use hypernetworks to generate adapter layers for multi-task language learning, while Polyhistor~\citep{liu2022polyhistor} adapts this for dense vision tasks. VL-Adapter~\citep{sung2022vl} introduces adapters for joint Vision-Language (VL) multi-task fine-tuning.

With the introduction of GPT-2~\citep{radford2019language} and the rise in parameter sizes and task variety, methods like Low-Rank Adaptation (LoRA)\citep{hu2022lora} have gained traction in both language and vision. These methods restrict weight updates to fixed low-dimensional subspaces, enhancing resource efficiency. Other approaches, such as SVFT\citep{lingam2024svft}, enforce sparse singular values, while DoRA~\citep{liu2024dora} decouples the magnitude and direction of weight update columns. In contrast, ReFT~\citep{wuandarora2024reft} uses low-rank approximation to adapt representations rather than weights.

Although most of these methods are not specifically designed for multi-task learning (MTL), MTLoRA~\citep{agiza2024mtlora} has applied LoRA within an MTL framework. By enabling both task-agnostic and task-specific adaptations along with joint inference, MTLoRA has achieved state-of-the-art results on dense vision tasks. However, these methods, while effective in preserving pre-trained knowledge, still impose subspace constraints that can degrade performance relative to single-task baselines.\\

\ourmethod\ takes a fundamentally different approach: instead of adding modules or modifying weights directly, we preserve the pre-trained singular vectors while enabling flexible adaptation via diffeomorphisms of singular values. This allows us to maintain both architectural efficiency and rich feature adaptability, without altering the input-output mapping learned during pre-training.
\\

\noindent\textbf{Multi-Task Learning Dynamics and Challenges.} 
A recent study of Transformer training dynamics reveals an incremental learning phenomenon where weight update ranks gradually increase during training~\cite {NEURIPS2023_4d69c1c0}. This finding challenges the effectiveness of fixed-rank adaptation methods in MTL settings. While MTLoRA~\citep{agiza2024mtlora} attempts to address this through task-agnostic and task-specific modules, it still operates within fixed low-rank constraints. Similarly, gradient interference mitigation techniques like PCGrad address conflicts between task updates but do not provide the flexibility needed for diverse task adaptations. In contrast, \ourmethod\ aligns with natural learning dynamics by preserving the full-rank structure through singular vector retention while enabling task-specific adaptations via neural diffeomorphisms~\citep{freifeld2015highly, freifeld2017transformations}, uniquely combining parameter efficiency with the adaptability needed for effective multi-task learning.
\\

\noindent\textbf{Diffeomorphisms in Neural Networks.}
A bijective mapping \( f:\gM \to \gN \) between two differentiable manifolds \(\gM\) and \(\gN\) is a \emph{diffeomorphism} if both \( f \) and its inverse \( f^{-1}:\gN \to \gM \) are differentiable. Learning diffeomorphisms poses computational challenges, as early methods were constrained by complex, infinite-dimensional spaces~\citep{NEURIPS2018_68148596}, and later Markov Chain Monte Carlo methods remained demanding~\citep{allassonniere2010construction, allassonniere2015bayesian, zhang2016bayesian}. \citet{freifeld2015highly, freifeld2017transformations} addressed these issues with Continuous Piecewise-Affine Based (CPAB) transformations, a finite-dimensional approach suited for precise 1D diffeomorphisms, ideal for neural activation functions. CPAB offers linear complexity, supports parallelization, and achieves sub-linear performance in practice~\cite{freifeld2017transformations}. Initially developed for alignment and regression, CPAB is now widely applied in neural networks. For instance, in spatial transformer layers with CPAB~\citep{detlefsen2018deep}, temporal alignment~\citep{martinez2022closed},  novel loss function for time-series analysis~\citep{weber2023regularization}, image animation~\citep{wang2024continuous}, geometry-guided fine-tuning~\citep{mantri2024rethinking} and trainable CPAB-based activation function~\citep{mantri2024digraf, Chelly2024DiTAC}.
\section{Preliminaries and Background}
\label{sec:background}
We combine two key components: the weights in ViTs, and the mathematical framework of CPAB diffeomorphisms. We now provide the background for understanding the coupling of these concepts in \ourmethod.

\subsection{Learned Weights in ViTs}
\label{sec:learned_weights_vits}
\begin{figure}[t]
    \centering
\includegraphics[width=1.0\linewidth]{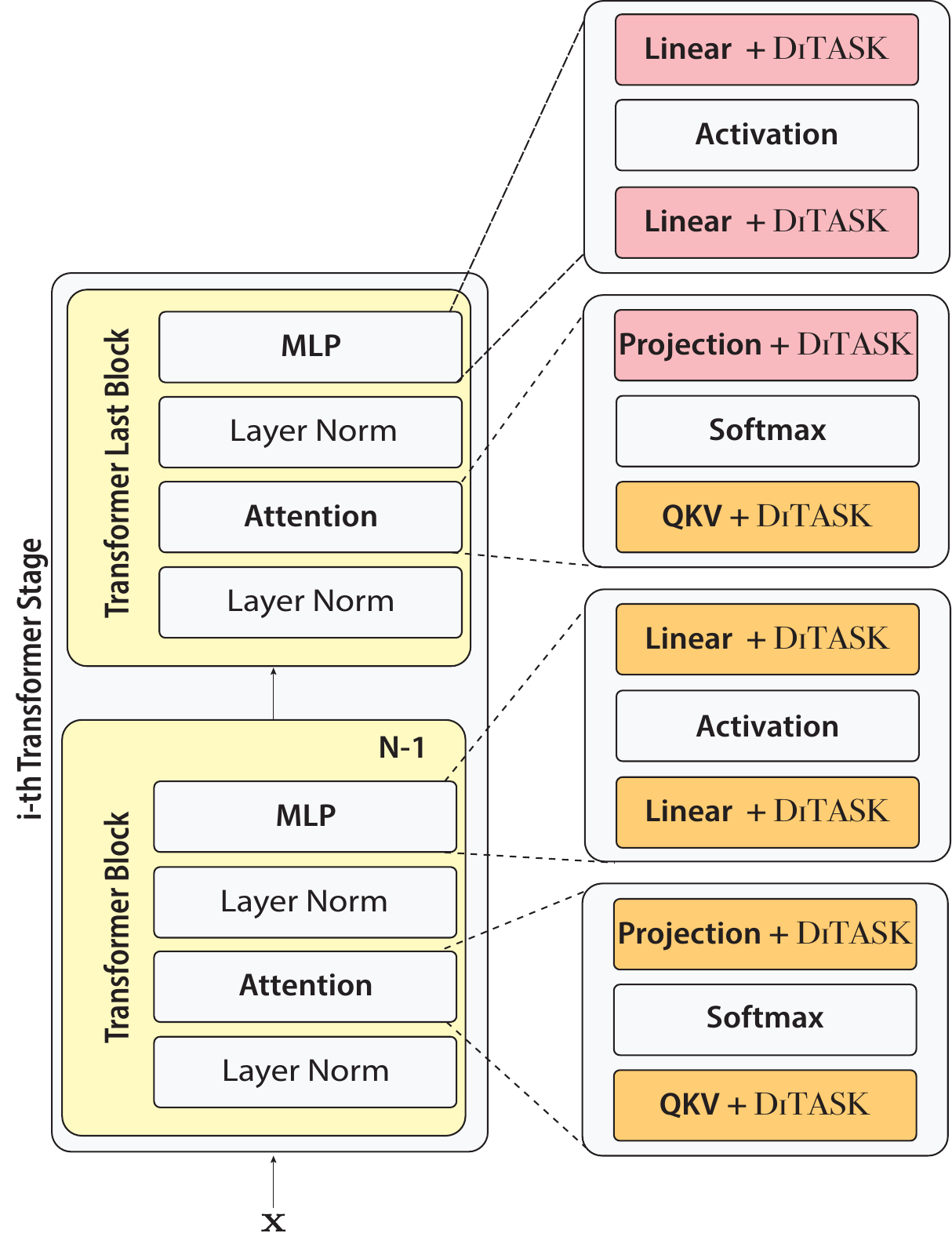}
    \caption{\textbf{\ourmethod\ in the $i$-th Swin Transformer stage for multi-task learning.} Task-agnostic modules are applied in all blocks except the last, which uses task-specific modules to capture task-dependent features.}

    \label{fig:mtlora}
\end{figure}
ViTs encode visual knowledge in their weight matrices, which can be analyzed through Singular Value Decomposition (SVD). Assume we have a weight matrix that transforms inputs from $c_1$  to $c_2$ channels, denoted by $\mathbf{W} \in \mathbb{R}^{c_2 \times c_1}$, then its SVD reads:
\begin{equation}
    \label{eq:svd_weight_matrix}
    \mathbf{W} = \mathbf{U} \mathbf{\Sigma} \mathbf{V}^\top,
\end{equation}
where $\mathbf{U} \in \mathbb{R}^{c_2 \times c_2}, \ \mathbf{\Sigma} \in \mathbb{R}^{c_2 \times c_1}, \ \mathbf{V} \in \mathbb{R}^{c_1 \times c_1}$. 
We refer to $\mathbf{U}$ as the \emph{left} singular vectors, $\mathbf{\Sigma}$ as the singular \emph{values}, and $\mathbf{V}$ as the \emph{right} singular vectors. Notably, $\mathbf{U}$ and $\mathbf{V}$ form orthonormal bases for the output and input spaces of the linear map represented by $\rmW$, respectively. These bases characterize how the network transforms visual features across layers and hidden dimensions. 
\\
\noindent\textbf{Problem Setup.} Throughout this paper, we assume  $K$ tasks $\mathcal{T} = \{T_1, \cdots, T_K\}$. Each task $T_k$ has an associated dataset $\mathcal{D}_k = \{(x_i^k, y_i^k)\}_{i=1}^{N_k}$ where $x_i^k \in \mathcal{X}$ and $y_i^k \in \mathcal{Y}_k$ represent the input-output pairs. The goal is to train a neural network to minimize a supervised loss between the prediction and ground-truth output. Following \cite{agiza2024mtlora}, we use a Swin Transformer~\cite{liu2021swin}, which processes input images by dividing them into non-overlapping patches and progressively merging patches to reduce spatial resolution while increasing feature dimension across stages. 
Formally, the Swin Transformer processes images through $L$ stages, with the $l$-th stage operating on feature maps of resolution $\frac{H}{2^l} \times \frac{W}{2^l}$ with $C_l$ channels. The key operation in each stage is a Window-based Multi-head Self-Attention (W-MSA):
\begin{equation}
    \text{W-MSA}(Q, K, V) = \text{Softmax}\left(\frac{QK^\top}{\sqrt{d}}\right)\,V,
\end{equation}
where $Q, K, V \in \sR^{M \times d}$ are query, key, and value matrices for windows of size $M$. The corresponding weight matrices $\rmW_l^{(q)}, \rmW_l^{(k)}, \rmW_l^{(v)} \in \sR^{d \times d}$ for these projections in layer $l$ are central to our \ourmethod{}, because in Section \ref{sec:method} we dynamically adjust their singular values for multi-task adaptations. The MTL objective minimizes a weighted combination of $K$ task losses:
\begin{equation}
\label{eqn:mtlloss}
    \min_{\Theta} \sum\limits_{k=1}^K \lambda_k \mathcal{L}_k(\Theta; \mathcal{D}_k),
\end{equation}
where $\mathcal{L}_k$ and $\lambda_k$ represent the loss function and weight for the $k$-th task, respectively. In \Cref{fig:mtlora}, we illustrate the overall architecture used in this paper.

\subsection{CPAB Diffeomorphisms for Weight Transformation}
\label{subsec:cpab}
As we show later in Section \ref{sec:method}, our \ourmethod{} modifies singular values $\mathbf{\Sigma}$ while preserving the structure of weight matrices. To achieve that, transformations that are smooth, invertible, and monotone, are required. Continuous Piecewise Affine-Based (CPAB) \cite{freifeld2015highly, freifeld2017transformations} diffeomorphisms satisfy these conditions  while being efficient in terms of parameters and computing diffeomorphisms. We now turn to define CPAB formally. 

\begin{definition}[CPAB Transformation]
    Given a closed interval $\Omega = [a, b]$ partitioned into $\mathcal{N}_{\mathcal{P}}$ intervals, a CPAB transformation is defined through a velocity field $v^\vtheta: \Omega \to \sR$ that is continuous and piecewise-affine within each interval. This velocity field, parameterized by $\vtheta \in \sR^{\gN_{\gP} - 1}$, yields a diffeomorphism $f^{\vtheta}$ of the form:
\begin{equation}
\label{eq:cpab}
    f^{\vtheta}(x) = x + \int_0^1 v^\vtheta(f^{\vtheta}(x, \tau))\, d\tau.
\end{equation}
\end{definition}
\noindent For practical implementation, we use the Python package from \citet{martinez2022closed}, and provide additional details in the supplementary material. 
\noindent
CPAB transformations are \emph{ideal for our settings}, shown in Section \ref{sec:method}, because they provide: (i) \textit{Low parameter overhead.} Only $\gO(\gN_\gP)$ parameters, where typically \(\gN_\gP \leq 128\); (ii) \textit{Differentiability.} Enables gradient-based optimization; (iii) \textit{Computational efficiency.} Computing \Cref{eq:cpab} is of linear complexity; (iv) \textit{Stability.} CPAB transformations are Lipschitz continuous \cite{freifeld2017transformations}.
\section{\ourmethod: Diffeomorphic Multi-Task Adaptation}
\label{sec:method}

We introduce \ourmethod\ (Diffeomorphic Multi-Task Adaptation), a novel approach for efficient multi-task fine-tuning. Our method is grounded in a fundamental analysis of how neural networks process information through their weight matrices. We draw insights from linear algebra to the workings of a linear layer of neural network in \Cref{subsec:motivation} relating it to SVD. In \Cref{subsec:ditask}, we formally introduce \ourmethod\ and elaborate on its properties that motivate our design choices in \Cref{subsec:theory}.

\subsection{Understanding Weight Matrices}
\label{subsec:motivation}
We now show that by considering the SVD of weight matrices, we can gain insights on the course of action of linear layers, that later, in Section \ref{subsec:ditask}, serves as a mathematically grounded motivation for the design of our \ourmethod. Specifically, we note that  a linear layer with weight matrix $\mathbf{W} \in \mathbb{R}^{c_2 \times c_1}$, and its SVD decomposition $\bf W = \bf U \bf \Sigma \bf V^\top$, can be viewed as a linear map $g: \mathbb{R}^{c_1} \rightarrow \mathbb{R}^{c_2}$, such that the following equality holds:
\begin{equation}
    g(\rvx) = \mathbf{W} \mathbf{x} = \sum\limits_{i=1}^{p} \sigma_i \, \langle \rvx, \rvv_i \rangle \rvu_i,
\end{equation}
where $\rmU = [\rvu_1, \ldots, \rvu_{c_2}]$, $\rmV = [\rvv_1, \ldots, \rvv_{c_1}]$, and $\mathbf{\Sigma} = diag([\sigma_1, \sigma_2, \ldots, \sigma_{p}])$, $p = \min(c_1, c_2)$. The basis vectors in $\rmU$ are the left singular vectors, $\bf V$ are right singular vectors, and $\mathbf{\Sigma}$ are the singular values of $\rmW$.

\noindent Specifically, the orientations encoded by the left and right singular vectors represent features present in the output and input spaces of the linear transformation $g$, respectively. Multi-task learning requires adapting these orientations \emph{jointly} for all tasks, while minimizing task interference to prevent performance degradation. Below, we describe how to achieve effective multi-task adaptation using diffeomorphisms, by preserving the singular vectors $\bf U, \ \bf V$ and modifying $\bf W$ in a principled manner.

\subsection{Adaptation with \ourmethod}
\label{subsec:ditask}
Given a learned weight matrix $\mathbf W = \mathbf{U} \mathbf{\Sigma} \mathbf{V}^\top \in \mathbb{R}^{c_2 \times c_1}$ from a ViT, we transform the singular values $\mathbf{\Sigma}$ by a learned neural diffeomorphism $f^{\vtheta}$ described in \Cref{eq:cpab}, to get the adapted weight matrix $\rmW_A$ as follows:
\begin{equation}
\label{eq:cpab_ditask}
     \rmW_A = \rmU \, \begin{bmatrix}
        f^{\vtheta}(\sigma_1) & 0 & \ldots & 0 \\
    0 & f^{\vtheta}(\sigma_2) & \ldots & 0 \\
     \vdots & \vdots & \ddots & \vdots \\
     0 & 0 & \ldots & f^{\vtheta}(\sigma_p)
    \end{bmatrix} \, \rmV^\top.
\end{equation}

\noindent During fine-tuning, we freeze $\rmW$ and learn only the parameters $\vtheta$ of the CPA velocity field. This allows \ourmethod\ to adapt to multiple tasks using $\gO(\gN_\gP)$ additional parameters (where typically  $\gN_\gP \leq 128$), which is significantly smaller than  methods with low-rank $r$ using $\gO(r(c_1 + c_2))$. We describe this process in the \Cref{fig:left}, and compare the computational cost of \ourmethod{} with other fine-tuning methods in the supplementary material.
\\
\noindent\textbf{Multi-Task Adaptation with \ourmethod.}  
To enable both multi and task-specific adaptation, we learn two sets of neural diffeomorphic transformations in every pre-trained weight matrix $\rmW$, shown in   \Cref{fig:overview_right}, that include: (i) \textit{Joint Adaptation} parameters, denoted by $\vtheta_j$, to allow task synergies which are crucial for MTL \citep{Huang_2024_CVPR,mtl_survey}; and (ii) \textit{Task-wise Adaptation} parameters, denoted by $\{\vtheta_k\}_{k=1}^K$ to enable learning nuances for each task.

\subsection{Properties of \ourmethod}
\label{subsec:theory}
In this section, we motivate our design choices and discuss the properties that make \ourmethod\ an effective and parameter-efficient fine-tuning method for MTL.
\\
\noindent\textbf{Feature Space Preservation.} As discussed in \Cref{subsec:motivation}, pre-trained weight matrices encode feature transformations through their singular vectors, where $\mathbf{U}$ and $\mathbf{V}$ form orthonormal bases for output and input spaces, respectively. These bases capture the orientation of patterns learned during pre-training, each weighted by its corresponding singular value. In contrast to traditional low-rank methods like LoRA, which constrain adaptations to fixed low-dimensional subspaces, potentially losing important features, our \ourmethod\ preserves these learned bases entirely. This preservation allows us to achieve superior multi-task performance with fewer parameters, as shown in \Cref{fig:teaser}. Specifically, while low-rank methods require increasing their rank (and thus parameters and departure from being low-rank) to match single-task performance, our \ourmethod\ maintains the learned feature space of pre-trained ViTs through a targeted and mathematically-ground singular value modulation.
\\
\noindent\textbf{Feature Relativity Preservation.} The order of singular values $\sigma_1 \geq \sigma_2 \geq \ldots \geq \sigma_p$ in weight matrices of ViTs encode the relative importance of features learned during pre-training on large-scale datasets like ImageNet-21k. By utilizing diffeomorphisms, which are smooth, invertible, and monotone, by definition, we preserve this ordering. That is, if $f^{\vtheta}$ is a diffeomorphism, then $f^{\vtheta}(\sigma_1) \geq f^{\vtheta}(\sigma_2) \geq ... \geq f^{\vtheta}(\sigma_p)$ in \Cref{eq:cpab_ditask}. Our choice of CPAB transformations implements this with only $\gN_\gP \leq 128$ parameters per weight matrix, compared to the $\gO(r(c_1 + c_2))$ parameters required by low-rank adaptation methods. Our experiments in \Cref{sec:exp} demonstrate that preserving feature hierarchy through diffeomorphic transformations enables effective multi-task generalization while maintaining parameter efficiency.

\section{Experiments}
\label{sec:exp}
    \begin{table*}[t]
      \centering
      \scriptsize
      \setlength{\tabcolsep}{2.5pt}
    \renewcommand{\arraystretch}{0.75}
    \scriptsize
    \caption{\textbf{Performance on PASCAL Context for Multi-Task Learning.} Comparison of \ourmethod{} with full fine-tuning, adapter-based, and fixed-rank methods across four dense prediction tasks: semantic segmentation, human part segmentation, saliency detection, and surface normal estimation. Metrics are \textsc{mIoU} ($\uparrow$) and \textsc{rmse} ($\downarrow$), with $\Delta m$ denoting relative improvement over single-task tuning. \ourmethod{} achieves strong performance with minimal trainable parameters and supports joint inference across tasks.}
      \vspace{-5pt}
    \small
      \begin{tabular}{l  cccc  cc  c }
        \toprule
        \multirow{2}{*}{\textbf{Method}} & \textbf{\textsc{SemSeg}} & \textbf{\textsc{Human Parts}} & \textbf{\textsc{Saliency}} & \textbf{\textsc{Normals}} & 
          \multirow{2}{*}{$\Delta m (\%) $} & \textbf{Trainable Swin} & \textbf{Single Inference}   \\
          & ($\textsc{mIoU} \uparrow$) & ($\textsc{mIoU} \uparrow$) & ($\textsc{mIoU} \uparrow$) & ($\textsc{rmse} \downarrow$) & & 
          \textbf{Parameters} (M) &
          \textbf{For All Tasks}   \\
        \midrule
        \textbf{\textsc{Full Fine-Tuning}} & \\
        Single Task & 67.21 & 61.93 & 62.35 & 17.97 & \(\phantom{-}0.00\) & 112.62   & $\times$ \\
        MTL - Dec. Only  & 65.09 & 53.48 & 57.46 & 20.69 & \(-\phantom{0}9.95\)  & \phantom{0}\phantom{0}0\phantom{0}\phantom{0}  & \checkmark  \\
        MTL - Full  &  67.56 & 60.24 & 65.21 & 16.64 & \(+\phantom{0}2.23\) &  \phantom{0}28.12    & \checkmark  \\
        \midrule
        \textbf{\textsc{Adapter-Based}} & \\
        VPT-shallow \cite{jia2022visual}  & 62.96 &  52.27 & 58.31 & 20.90 & \(-11.18\)  & \phantom{0}\phantom{0}0.63  & $\times$  \\
        VPT-deep \cite{jia2022visual}  & 64.35 & 52.54 & 58.15 & 21.07 & \(-10.85\) & \phantom{0}\phantom{0}1.49  & $\times$ \\
        Compactor++ \cite{karimi2021compacter}  & 67.26 & 55.69 & 59.47 & 19.54 & \(-\phantom{0}5.84\) & \phantom{0}\phantom{0}0.72    & $\times$ \\
        Bitfit \cite{zaken2021bitfit} & 68.57 & 55.99 & 60.64 & 19.42 & \(-\phantom{0}4.60\) & \phantom{0}\phantom{0}0.91   & $\times$  \\
        Compactor \cite{karimi2021compacter} & 68.08 & 56.41 & 60.08 & 19.22 & \(-\phantom{0}4.55\) & \phantom{0}\phantom{0}0.84   & $\times$  \\
        
        Adapter \cite{he2021towards} &  69.21 &  57.38 & 61.28 & 18.83  & \(-\phantom{0}2.71\) & \phantom{0}\phantom{0}9.26  & $\times$ \\
        VL-Adapter \cite{sung2022vl} & 70.21 & 59.15 & 62.29 & 19.26 & \(-\phantom{0}1.83\)  & \phantom{0}\phantom{0}2.80  & $\times$ \\
        Polyhistor \cite{liu2022polyhistor} & 70.87  & 59.15  & 65.54 &  17.77 & \(+\phantom{0}2.34\)  & \phantom{0}\phantom{0}7.02  & $\times$ \\
        HyperFormer \cite{mahabadi2021parameter}  & 71.43 & 60.73 & 65.54 &  17.77 & \(+\phantom{0}2.64\)  & \phantom{0}70.83  & $\times$ \\
        \midrule
        \textbf{\textsc{Fixed-Rank}} & \\
        MTL - DoRA~\citep{liu2024dora} & 52.36 & 50.82 & 63.53 & 18.32 & \(-10.02\) & \phantom{0}\phantom{0}6.40 & \checkmark\\
        MTL - ReFT~\citep{wuandarora2024reft} & 68.75 & 56.49 & 58.76 & 20.54 & \(-\phantom{0}6.63\) & \phantom{0}16.16 & \checkmark \\
        MTL - SVFT~\citep{lingam2024svft} & 64.44 & 55.94 & 63.03 & 17.86 & \(-\phantom{0}3.02\) & \phantom{0}\phantom{0}3.83 & \checkmark\\
    LoRA \cite{hu2022lora} & 70.12 & 57.73 & 61.90 & 18.96 & \(-\phantom{0}2.17\) & \phantom{0}\phantom{0}0.93   & $\times$ \\

        MTLoRA~\citep{agiza2024mtlora} (\(r = 16\)) & 68.19 & 58.99 & 64.48 & 17.03 & \(+\phantom{0}1.35\) & \phantom{0}\phantom{0}3.01  & \textbf{\checkmark} \\
        MTLoRA~\citep{agiza2024mtlora} ($r = 32$) & 67.74 & 59.46 & 64.90 & 16.59 & \(+\phantom{0}2.16\) & \phantom{0}\phantom{0}4.14  & \textbf{\checkmark} \\
        MTLoRA~\citep{agiza2024mtlora} \revision{($r = 64$)} & 67.90 & 59.84 & 65.40 & 16.60 & \(+\phantom{0}2.55\) & \phantom{0}\phantom{0}6.40  & \textbf{\checkmark} \\
        \midrule
        Single Task - \ourmethod & \textbf{72.20} & \textbf{62.33} & \textbf{65.70} & \textbf{16.55} & \(\mathbf{+\phantom{0}5.33}\) & \phantom{0}\phantom{0}1.60 ($\times$4) & $\times$\\
        MTL - \ourmethod\ (Ours) & 69.66 & 62.02 & 65.00 & 17.10 & \(\mathbf{+\phantom{0}3.22}\) & \phantom{0}\phantom{0}1.61 & \checkmark\\
        
        \bottomrule
      \end{tabular}
      \label{tab:results}
    \end{table*}

We evaluate \ourmethod{} on multiple MTL benchmarks and compare it with other adaptation methods. \Cref{subsec:expsetup} details the experimental setup and baselines,   \Cref{subsec:results} discusses our main findings, and \Cref{sec:ablation} presents few ablation studies of \ourmethod.
 We seek to address three key research questions: 
\begin{itemize}
    \item (RQ1) Can a selective modification of singular values while preserving singular vectors bridge the performance gap between decoder-only and full fine-tuning?
    
    \item (RQ2) Does full-rank preservation by singular value modulation outperform low-rank adaptation methods?
    
    \item (RQ3) How does \ourmethod\ compare to other parameter-efficient adaptation methods?

\end{itemize}

\subsection{Experimental Setup}
\label{subsec:expsetup}
\noindent\textbf{Datasets and Tasks.} We evaluate \ourmethod\ on the PASCAL-MTL dataset~\citep{pascal} (PASCAL-Context split) following the literature on multi-task dense prediction~\citep{vandenhende2020mti, xu2018pad, ye2022inverted}. This dataset includes 4,998 training and 5,105 validation images with annotations for semantic segmentation (21 classes), human part segmentation (7 classes), surface normal estimation, and saliency detection. Additional experiments on the  NYUD dataset~\citep{silberman2012indoor} are provided in \Cref{tab:nyud}.
\\
\noindent\textbf{Implementation Details.} We use the Swin-Tiny~\citep{liu2021swin} architecture, pre-trained on ImageNet-21k \cite{russakovsky2015imagenet} as the backbone encoder, similar to \cite{agiza2024mtlora}. To ensure a fair comparison, we incorporate the HRNet~\citep{hr_net} decoder, which accounts for 6\% of the total parameters, as in previous works \cite{agiza2024mtlora,vandenhende2020mti}.  Following MTINet~\citep{vandenhende2020mti}, we apply cross-entropy loss for segmentation tasks, $\ell_1$ loss for surface normals, and balanced cross-entropy \cite{cui2019classbalancedlossbasedeffective} for saliency detection, with same task weights $\{\lambda_k\}_{k=1}^K$ used in MTINet. In all experiments, we follow the evaluation protocol from \citet{agiza2024mtlora}.

\noindent\textbf{Baselines.} We benchmark our \ourmethod{} with three categories of methods:
\begin{table*}[t]
  \centering
  \scriptsize
   \setlength{\tabcolsep}{7.5pt}
  \caption{Effect of Backbone Size  -- Relative improvement increases with increasing capacity of the backbone}
  \vspace{-5pt}
\small
  \begin{tabular}{l  cccc  c  c  }
    \toprule
    \multirow{2}{*}{\textbf{Method}} & \textbf{\textsc{SemSeg}} & \textbf{\textsc{Human Parts}} & \textbf{\textsc{Saliency}} & \textbf{\textsc{Normals}} &
      \multirow{2}{*}{$\Delta m (\%) $} & \textbf{Trainable Swin} \\
      & ($\textsc{mIoU} \uparrow$) & ($\textsc{mIoU} \uparrow$) & ($\textsc{mIoU} \uparrow$) & ($\textsc{rmse} \downarrow$) & & 
      \textbf{Parameters} (M)    \\
    \midrule
    \ourmethod\ + Swin-Tiny & 69.66 & 62.02 & \textbf{65.00} & 17.10 & \(+3.22\) & 1.61 \\
    \ourmethod\ + Swin-Small & 74.49 & 63.20 & 64.58 & 17.58 & \(+4.68\) & 1.66 \\
    \ourmethod\ + Swin-Base & 75.86 & 65.97 & 64.18 & 17.29 & \(+6.52\) & 3.14 \\
    \ourmethod\ + Swin-Large & \textbf{76.23} & \textbf{67.53} & 64.07 & \textbf{16.90} & \(\mathbf{+7.79}\) & 7.13 \\
    \bottomrule
  \end{tabular}
  \label{tab:backbone}
\end{table*}
\begin{itemize}
    \item \textit{Full Fine-Tuning}: Methods that modify all weight matrices (single-task, MTL decoder, full MTL)
    \item \textit{Adapter-based}: Methods that add task-specific layers while preserving main weights (Adapter~\citep{houlsby2019parameterefficienttransferlearningnlp}, BitFit~\citep{zaken2021bitfit}, VPT~\citep{jia2022visual}, Compactor~\citep{karimi2021compacter}, VL-Adapter~\citep{sung2022vl}, HyperFormer~\citep{mahabadi2021parameter}, Polyhistor~\citep{liu2022polyhistor})
    \item \textit{Fixed-Rank}: Methods that constrain updates to low-rank subspaces (LoRA~\citep{hu2022lora}, DoRA~\citep{liu2024dora}, SVFT~\citep{lingam2024svft}, ReFT~\citep{wuandarora2024reft}, MTLoRA~\citep{agiza2024mtlora}\,)
\end{itemize}

\subsection{Results and Discussion}
  \label{subsec:results}
\Cref{tab:results} presents our main results, addressing (RQ1)--(RQ3) on the effectiveness of \ourmethod{} for multi-task learning (MTL).

\begin{figure}[h]
    \centering
    \includegraphics[width=1.0\linewidth]{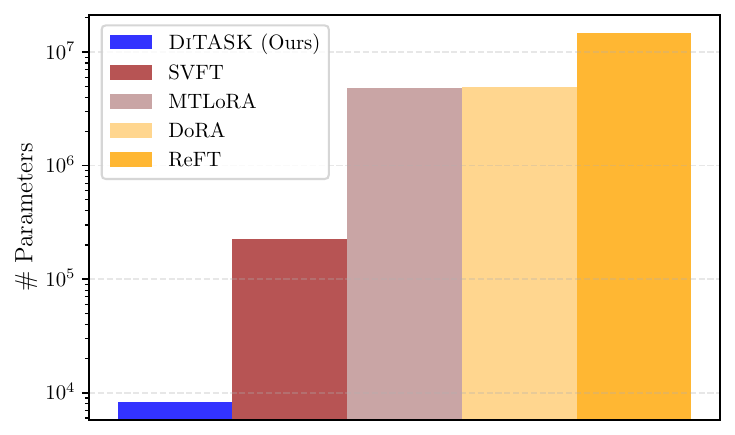}
    \caption{Comparison of adaptation parameter budget of the shared encoder, excluding LayerNorm parameters.}
    \label{fig:parameterallocation}
\end{figure}

\noindent\textbf{Bridging the Efficiency–Performance Gap.} \ourmethod{} achieves a \(\mathbf{+3.22\%}\) relative improvement over single-task baselines using only 1.61M trainable encoder parameters. This outperforms both decoder-only tuning (\(-9.95\%\)) and full fine-tuning (\(+2.23\%\)), demonstrating that modulating singular values while preserving pre-trained feature directions enables an effective balance between parameter efficiency and performance.
\\
\noindent\textbf{Comparison with Fixed Low-Rank Methods.}  
\ourmethod{} outperforms the state-of-the-art MTLoRA by \(26.27\%\) in \(\Delta m\), using \(\mathbf{4\times}\) fewer parameters (\Cref{fig:parameterallocation}). Most fixed-rank methods yield negative \(\Delta m\) due to task interference from shared low-rank subspaces. By modulating singular values independently per task while preserving full-rank structure, \ourmethod{} mitigates this interference. Similar trends are observed on NYUD (\Cref{tab:nyud}).
\\
\noindent\textbf{Parameter Efficiency and Task Performance.}
\ourmethod{} is the only parameter-efficient method to consistently improve over single-task baselines, showing that modulating singular values while preserving pre-trained directions effectively balances efficiency and performance.

\subsection{Ablation Study}
\label{sec:ablation}
We analyze key components of \ourmethod\ to understand their contributions to multi-task performance.
\\
\noindent\textbf{Computational Efficiency.}
\ourmethod\ uses only 8.5K adaptation parameters, compared to MTLoRA's 4.86M, while achieving 22\% faster batch processing (27.31s vs 35.20s) and reduced memory usage (see \Cref{tab:runtime}). The efficiency gains stem from our compact CPAB parameterization of singular value transformations.
\\
\noindent\textbf{Joint vs Task-Wise Singular Value Modulation.}
We evaluate three configurations of singular value modulation, shown in  \Cref{fig:ab1}. Joint Adaptation alone, with shared modulation across tasks, achieves an improvement of 2.77\%. Task-Wise Adaptation alone, which modulates singular values independently for each task, yields 2.85\%. The Combined approach, \ourmethod, incorporates both strategies and reaches the highest improvement at 3.22\%, supporting the use of both shared and task-specific transformations.
\\
\noindent\textbf{Model Scaling Analysis.}
The effectiveness of \ourmethod{} scales with model capacity (\Cref{tab:backbone}). This scaling effect indicates that larger models provide richer singular value spaces, enabling more effective task-specific adaptation.
\begin{table}[t]
    \centering
    \scriptsize
    \caption{Comparison of computational efficiency between \ourmethod{} and MTLoRA. \ourmethod{} reduces batch runtime by \(\approx 22\%\), lowers memory usage, and uses \(500\times\) fewer adaptation parameters.}

    \label{tab:runtime}
    \begin{adjustbox}{width=\columnwidth}
    \begin{tabular}{lccc}
      \toprule
      Method & Batch Runtime & Max GPU Mem  & \# Adaptation \\
      & (seconds) & (Megabytes) & Params\\
      \midrule
        MTLoRA & 35.20$\pm$1.05 & 23509 & 4.86\textsc{m}\\
        \ourmethod & 27.31$\pm$5.29 & 22906 & 8.5$\textsc{k}$\\
      \bottomrule
    \end{tabular}
  \end{adjustbox}
\end{table}
\noindent\textbf{Qualitative Comparison.} In \Cref{fig:viz}, we present a visual comparison between MTLoRA and our \ourmethod\ on the semantic segmentation task from PASCAL MTL. \ourmethod\ demonstrates tighter and clearer boundaries, more consistent segmentation, and improved class separation. For example, in the dimly lit sofa scene in the second column, MTLoRA struggles to correctly predict the sofa boundary, while \ourmethod\ offers a better prediction. 

\begin{table}[t]
    \centering
    \begin{adjustbox}{width=\columnwidth}
  \begin{tabular}{l cccc}
    \toprule
         Method & \textbf{\textsc{SemSeg}} & \textbf{\textsc{Depth}} & \multirow{2}{*}{${\Delta m (\%)}$} & \textbf{Trainable Params} \\
         & $(\textsc{mIoU} \uparrow)$ & $(\textsc{rmse} \downarrow)$ & & (in M)\\
         \midrule
         Single Task & 33.18 & 0.667 & 0 & 112.62 \\
         MTL - Dec. Only & 28.37 & 0.832 & -19.61 & 1.00\\
         MTL - Full & 35.29 & 0.734 & -1.84 & 28.5\\
         \midrule
         MTLoRA & 37.18 & 0.635 & +8.42 & 6.26 \\
         \midrule
         Single Task - \ourmethod & \textbf{44.01} & 0.644 & +18.04 & 1.61\\
         \ourmethod & 43.85 & \textbf{0.606} & +\textbf{20.65} & 1.61\\
         \bottomrule
    \end{tabular}
  \end{adjustbox}
    
    \caption{Comparison of \ourmethod{} and MTLoRA on the NYUD~\citep{silberman2012indoor} multi-task dataset for semantic segmentation (\textsc{mIoU}) and depth estimation (\textsc{rmse}).}

    \label{tab:nyud}
\end{table}
\section{Conclusion}
\label{sec:conclusion}

\begin{table}[t]
\centering
\setlength{\extrarowheight}{1em}
\newcommand{\imwidth}{0.067\textwidth}
\setlength{\tabcolsep}{1pt}
\begin{tabular}{>{\centering\arraybackslash}m{2cm} >{\centering\arraybackslash}m{\imwidth} >{\centering\arraybackslash}m{\imwidth} >{\centering\arraybackslash}m{\imwidth} >{\centering\arraybackslash}m{\imwidth} >{\centering\arraybackslash}m{\imwidth}}
{Input Image} & \includegraphics[valign=c,width=\imwidth]{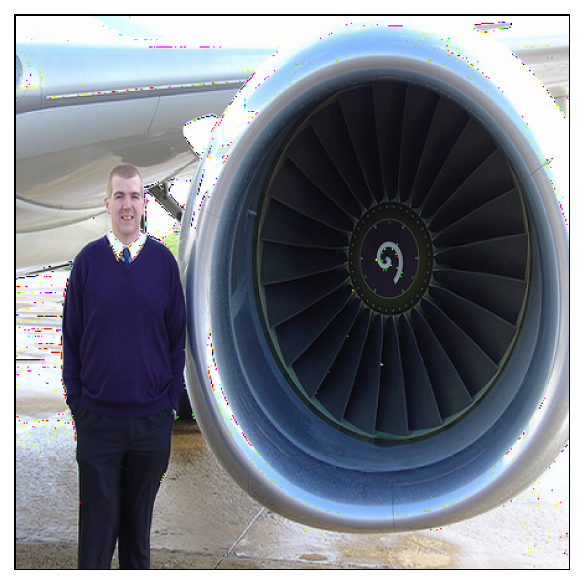} & \includegraphics[valign=c,width=\imwidth]{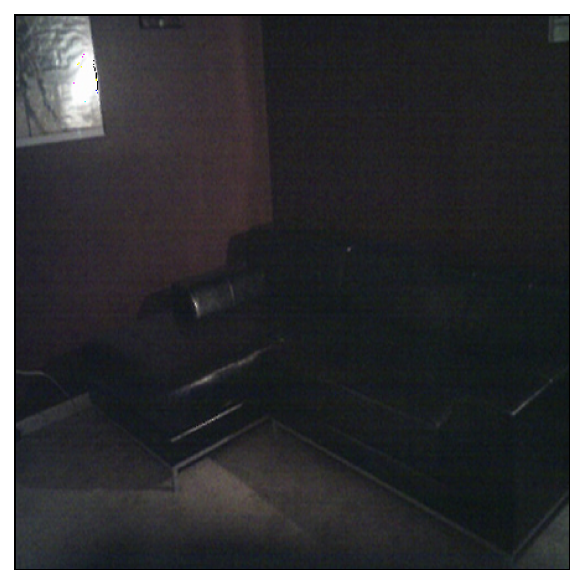} & \includegraphics[valign=c,width=\imwidth]{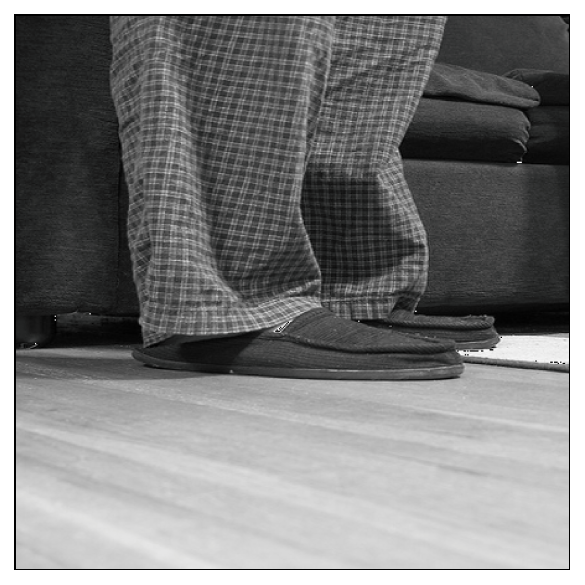} & \includegraphics[valign=c,width=\imwidth]{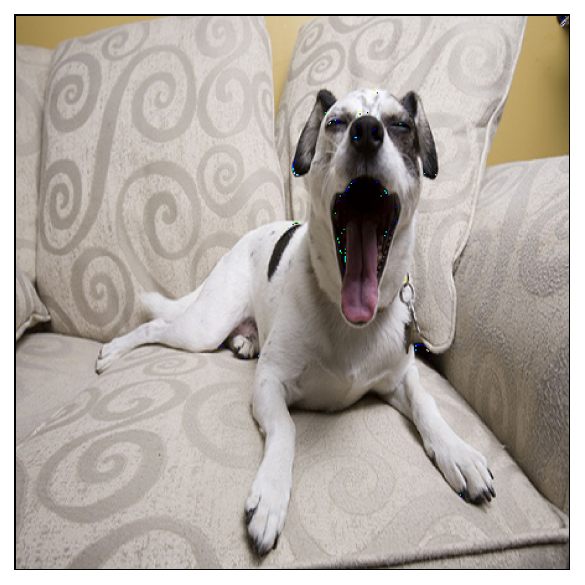} & \includegraphics[valign=c,width=\imwidth]{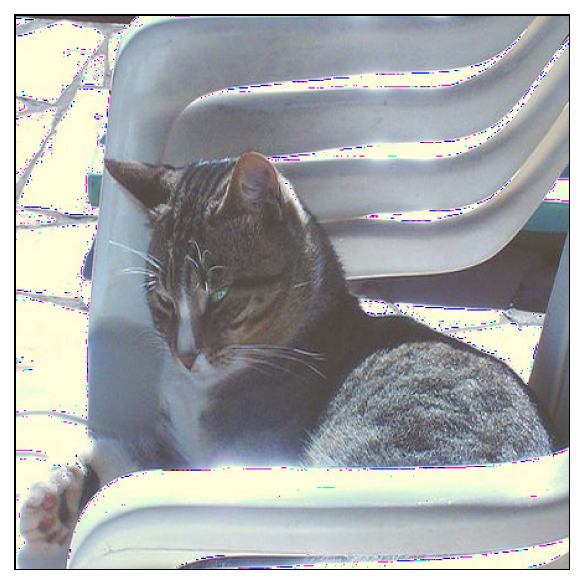} \\
{MTLoRA} & \includegraphics[valign=c,width=\imwidth]{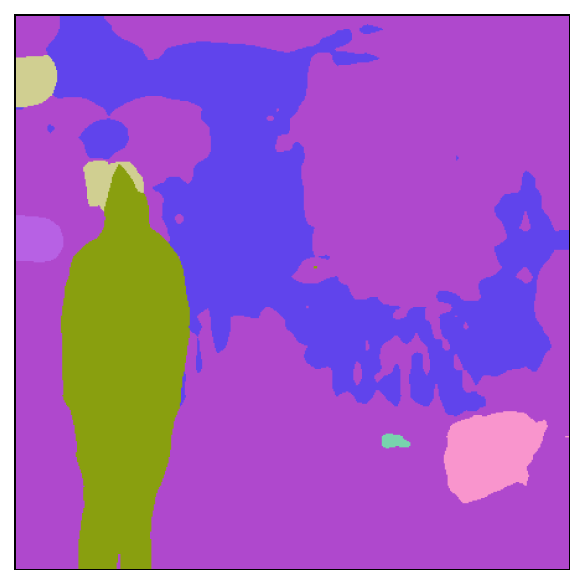} & \includegraphics[valign=c,width=\imwidth]{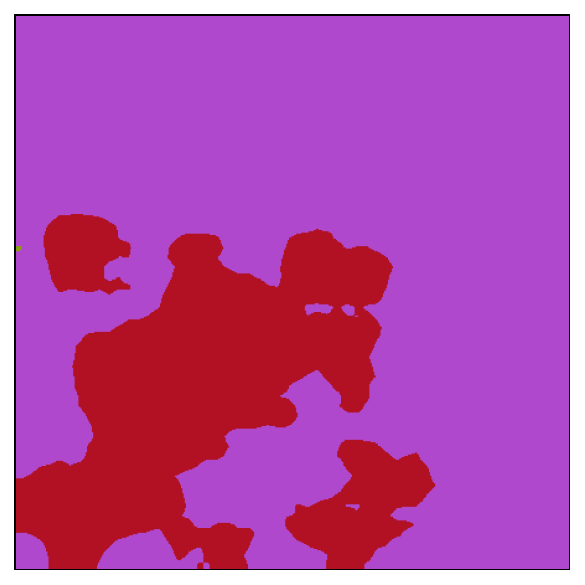} & \includegraphics[valign=c,width=\imwidth]{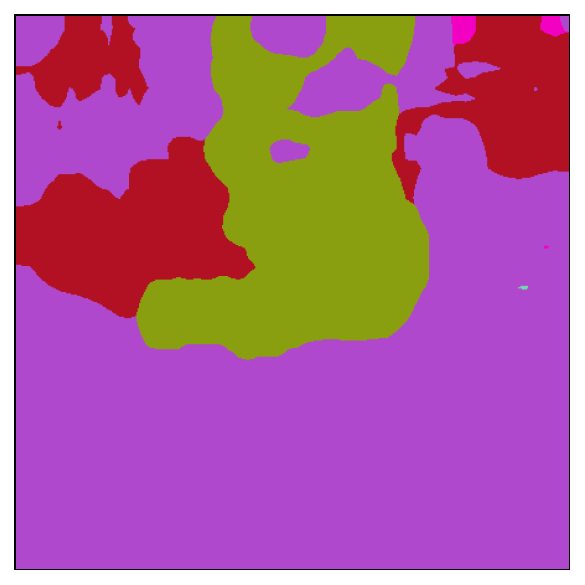} & \includegraphics[valign=c,width=\imwidth]{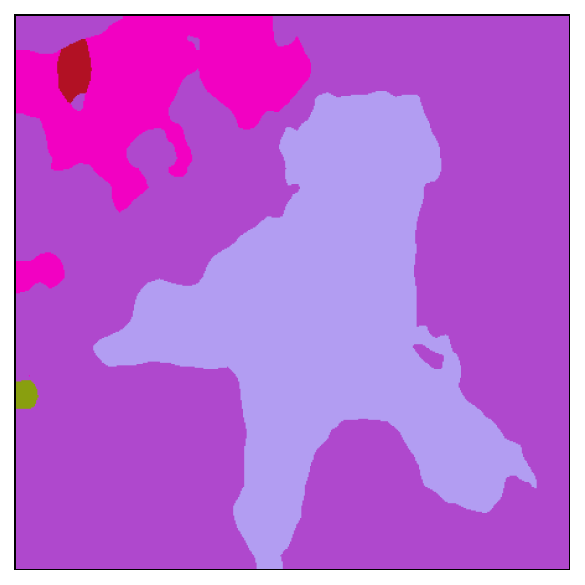} & \includegraphics[valign=c,width=\imwidth]{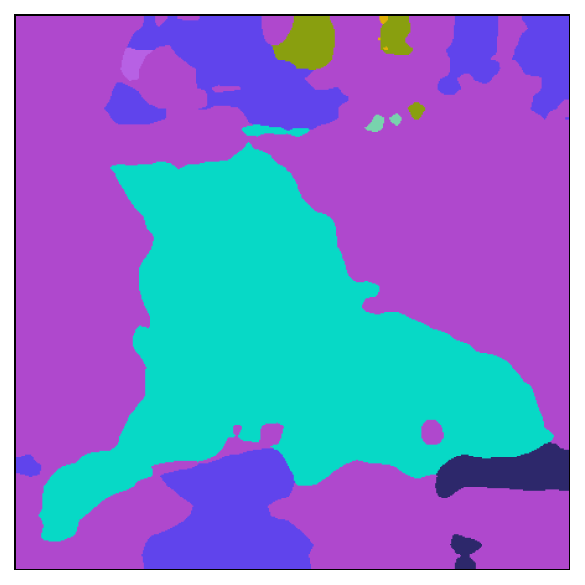} \\
{\textsc{DiTASK}} & \includegraphics[valign=c,width=\imwidth]{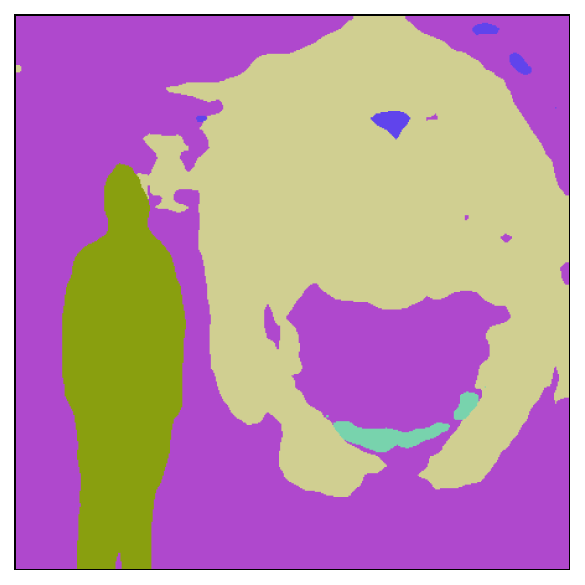} & \includegraphics[valign=c,width=\imwidth]{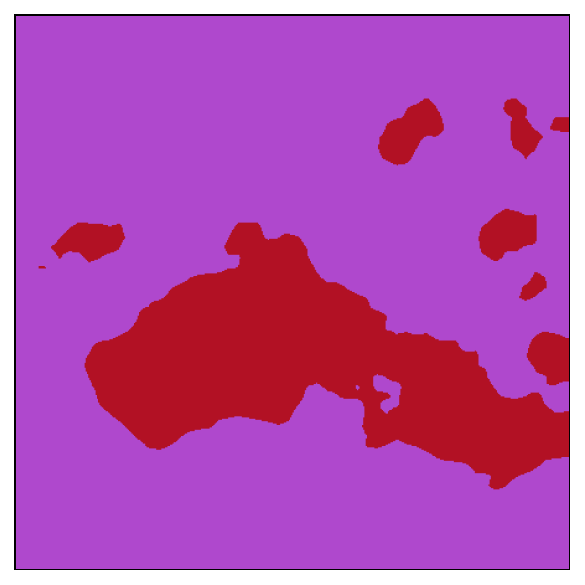} & \includegraphics[valign=c,width=\imwidth]{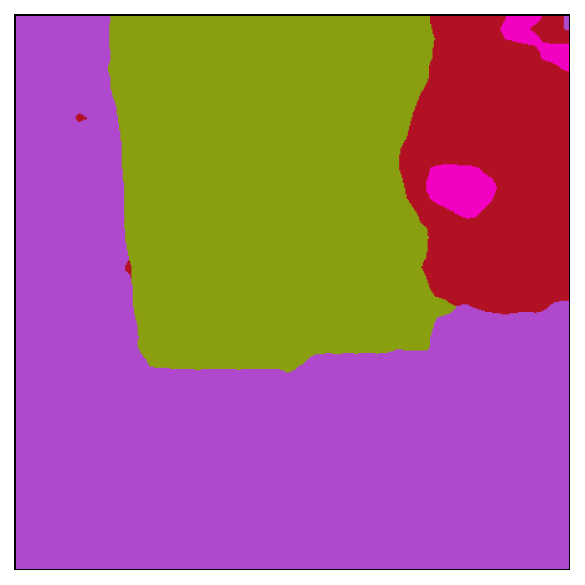} & \includegraphics[valign=c,width=\imwidth]{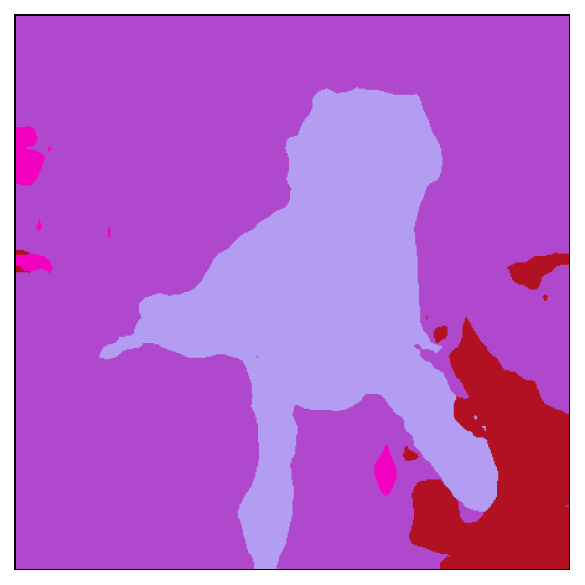} & \includegraphics[valign=c,width=\imwidth]{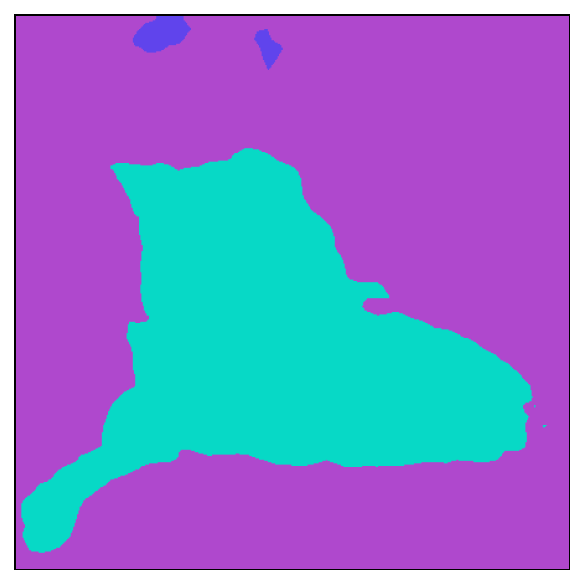} \\
{Ground Truth} & \includegraphics[valign=c,width=\imwidth]{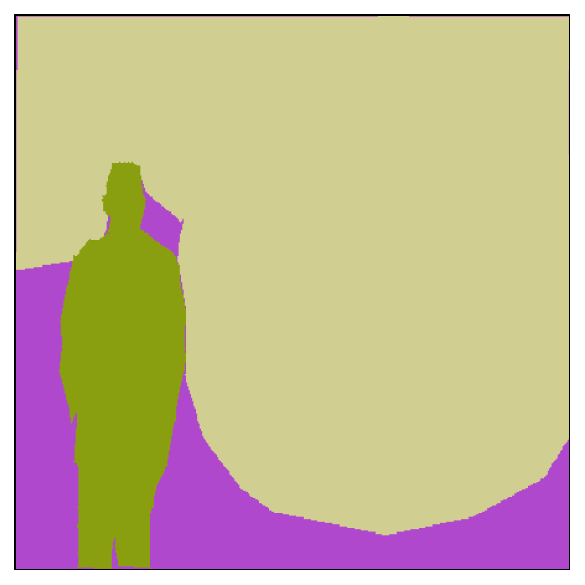} & \includegraphics[valign=c,width=\imwidth]{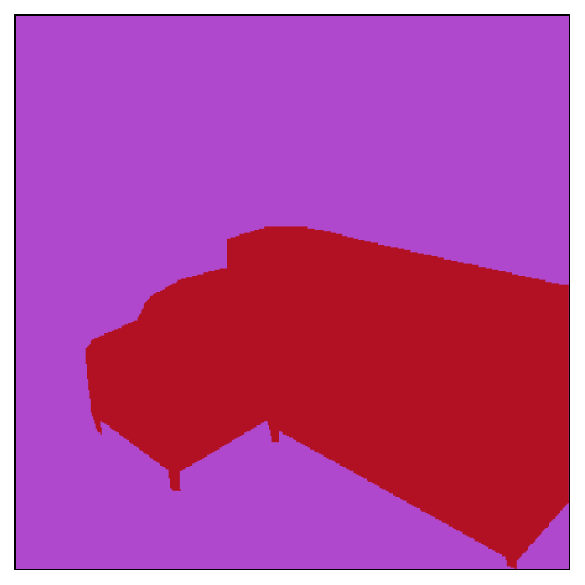} & \includegraphics[valign=c,width=\imwidth]{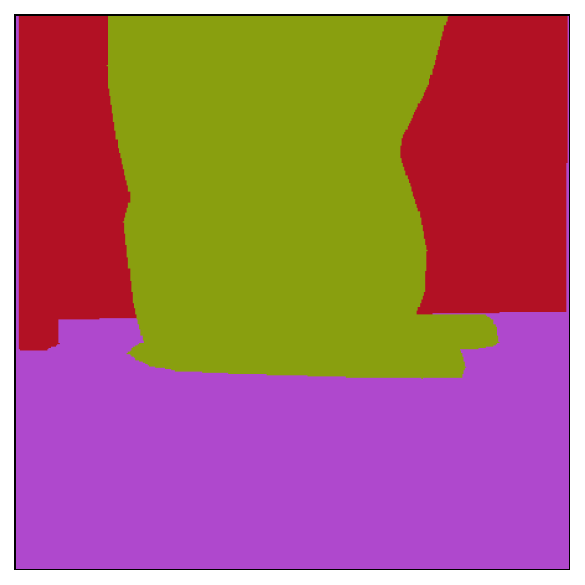} & \includegraphics[valign=c,width=\imwidth]{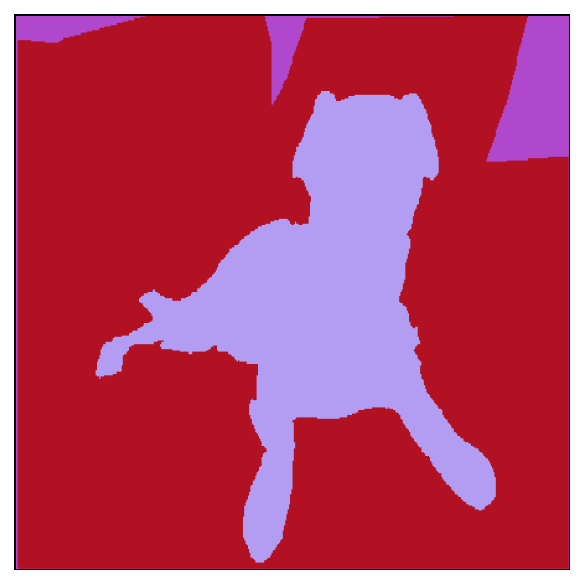} & \includegraphics[valign=c,width=\imwidth]{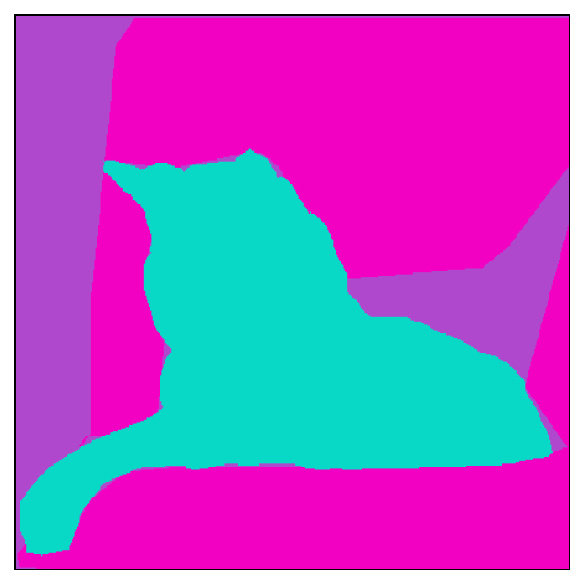} \\
\end{tabular}
\captionof{figure}{Semantic segmentation predictions on PASCAL MTL using MTLoRA and \ourmethod{}.
}
\label{fig:viz}
\end{table}

In this paper, we introduce \ourmethod, a novel MTL fine-tuning approach that leverages neural diffeomorphisms for singular value adaptation while preserving the structure of pre-trained representations. We observe that existing methods, which restrict weight updates to fixed low-rank subspaces for resource efficiency, often suffer from multi-task performance degradation relative to single-task baselines. We conjecture that the singular vectors of pre-trained weight matrices capture rich features, and preserving them during MTL adaptation enhances performance. By using neural diffeomorphisms, \ourmethod\ maintains feature space and relational structure while requiring less memory than other low-rank adaptation methods. We evaluate \ourmethod\ on the PASCAL MTL and NYUD MTL datasets, with extensive ablation studies demonstrating its effectiveness. Our results show that \ourmethod\ achieves state-of-the-art MTL performance with 75\% fewer trainable parameters, underscoring the importance of preserving singular vectors in pre-trained weights.

{
    \bibliography{main}
}
 \appendix
\clearpage
\setcounter{page}{1}
\maketitlesupplementary
\begin{figure}[t]
  \centering
\includegraphics[width=1.0\linewidth]{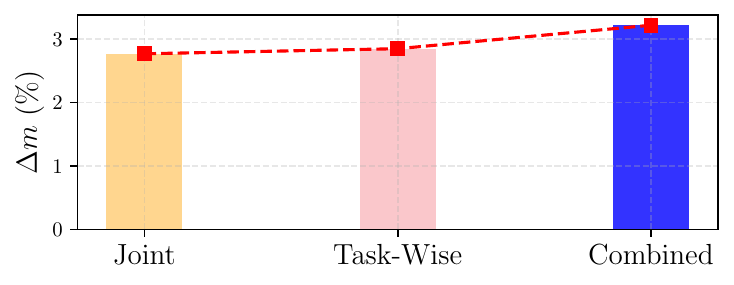} 
  \caption{Effect of task-specific and task-agnostic components}
  \label{fig:ab1}
\end{figure}
\section{\ourmethod\ Adaptation Process}
In this section, we present the pseudocode for our \ourmethod. We repeat this adaptation process in every training iteration for every transformer stage in the encoder, as shown in \Cref{fig:mtlora}.
\begin{quote} 
\noindent\texttt{\underline{\ourmethod($\rmW, \vtheta_j, \{\vtheta_k\}_{k=1}^K, f, \rvx, \{\rvx_k\}_{k=1}^K$)}}
\begin{steps}
    \item Compute the Singular Value Decomposition (SVD) of \( \rmW \):
    \begin{steps}
        \item \( \rmW = \rmU \mathbf{\Sigma} \rmV^\top \), where:
         \( \rmU \in \mathbb{R}^{c_2 \times c_2}, \mathbf{\Sigma} = \text{diag}(\sigma_1, \ldots, \sigma_p), \rmV \in \mathbb{R}^{c_1 \times c_1}. \)
    \end{steps}
    \item[\textit{//}] \textit{Joint Adaptation}
    \item  \( \mathbf{\Sigma}_J = \text{diag}(f^{\vtheta_j}(\sigma_1), \cdots, f^{\vtheta_j}(\sigma_p)). \)
    \item Construct \( \rmW_J = \rmU \mathbf{\Sigma}_J \rmV^\top. \)
    \item $\rvh = \rmW_J \, \rvx$
    \item[\textit{//}] \textit{Task-Specific Adaptation}
    \item For $k = 1, \cdots, K$ do
    \begin{steps}
            \item $\rvx_k = \rvx$ if not last block, else $\rvx_k$
            \item \( \mathbf{\Sigma}_k = \text{diag}(f^{\vtheta_k}(\sigma_1), \cdots, f^{\vtheta_k}(\sigma_p)). \)
            \item Construct \( \rmW_k = \rmU \mathbf{\Sigma}_k \rmV^\top. \)
            \item $\rvh_k = \rmW_k\,\rvx_k$
    \end{steps}
    \item \textbf{Return} $\rvh, \{\rvh_k\}_{k=1}^K$
\end{steps}
\end{quote}

\section{Gradient Analysis}
We analyze the memory requirements for low-rank adaptation methods, such as LoRA, and compare them with \ourmethod\ in terms of gradient storage.

LoRA adapts a pre-trained weight matrix $\rmW \in \sR^{c_2 \times c_1}$ using two learnable low rank matrices $\rmB \in \sR^{c_2 \times r}$ and $\rmA \in \sR^{r \times c_1}$. During backpropagation, gradients need to be stored for both $\rmB, \rmA$, and the input $\rvx$, resulting in a memory requirement that scales with $rc_2(c_2+c_1) + c_1 c_2$. This scaling depends directly on the rank $r$, which can make LoRA memory-intensive when $r$ is large or when the dimensions $c_1$ and $c_2$ are significant.

In contrast, \ourmethod\ leverages the singular value decomposition (SVD) of $\rmW = \rmU \mathbf{\Sigma}\rmV^\top$, where $\mathbf{\Sigma} = \text{diag}(\sigma_1, \cdots, \sigma_p), p=\min(c_1, c_2)$. \ourmethod\ adapts $\rmW$ by learning transformations on the singular values $\mathbf{\Sigma}$ using a small set of parameters $\vtheta$. This reduces the gradient storage requirement to $\gN_\gP + p + c_1 c_2$, where $\gN_\gP$ is the number of intervals over which the CPA velocity field is defined. Unlike LoRA, \ourmethod\ avoids gradients tied to low-rank matrices, significantly reducing memory usage for tasks with high-rank requirements or large input dimensions.

By operating directly on the singular values, \ourmethod\ achieves a more memory-efficient adaptation strategy while retaining the ability to make task-specific updates. This efficiency makes it particularly advantageous for large-scale models.

\section{Experimental and Implementation Details}
\label{sec:appendix:hyper}
\noindent\textbf{Training.} We train using AdamW~\citep{loshchilov2019decoupledweightdecayregularization} optimizer with StepLR scheduler for 300 epochs on 8 NVIDIA A6000 GPUs (batch size 64 per GPU).

\noindent\textbf{Code.} Our implementation closely follows the codebase of MTLoRA~\citep{agiza2024mtlora} (MIT License), which we modified for our requirements. We refactored the code to allow distributed training.

\noindent\textbf{Hyperparameters.} The hyperparameters specific to \ourmethod\ are the tessellation size $\gN_\gP$ for each joint and task-wise transformations. In all our experiments, we perform a hyperparameter grid search using Weights \& Biases framework~\cite {wandb}. All our experiments were performed on a single node with 8 NVIDIA A6000 Ada GPUs using distributed data-parallel (DDP) training
\begin{itemize}
    \item \textbf{PASCAL MTL:} We used the tessellation size of CPAB transformations in $\{16, 32, 64, 128\}$, dropout in $\{0.0, 0.05, 0.5\}$, learning rate in $\{0.005, 0.0005, 0.00005\}$, warmup epochs for StepLR in $\{20, 30, 40\}$, weight decay in $\{0.05, 0.005, 0.0005, 0.00005\}$ and a total training epochs of 300 with a batch size of 64 per GPU.
    \item \textbf{NYUD:} We used the tessellation size of CPAB transformations in $\{16, 32, 64, 128\}$, dropout in $\{0.0, 0.05, 0.5\}$, learning rate in $\{0.005, 0.0005\}$, warmup epochs for StepLR in $\{10, 20, 30\}$, weight decay in $\{0.05, 0.005, 0.0005, 0.00005\}$ and a total training epochs of 100 with a batch size of 64 per GPU.
\end{itemize}

\noindent\textbf{Evaluation Metrics.} We follow the evaluation protocol of   MTLoRA \cite{agiza2024mtlora}:
\begin{itemize}
    \item Task-specific metrics: \textsc{mIoU} for segmentation tasks and \textsc{rmse} for surface normals and depth estimation.
    \item Average relative improvement across tasks:
    \begin{equation}
    \Delta m = \frac{1}{K} \sum\limits_{k=1}^K (-1)^{l_k}\,\frac{(M_k - M_{st, k})}{M_{st, k}},
    \end{equation}
    where $M_k$ is the performance on task $k$, $M_{st, k}$ is the single-task baseline. $l_k=1$ for metrics where lower is better, and 0 otherwise.
\end{itemize}

\section{Additional Ablations}
\begin{figure}[t]
  \centering\includegraphics[width=0.9\linewidth]{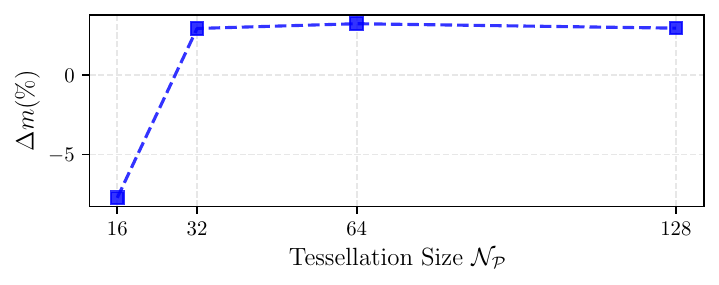}
  \caption{Effect of tessellation size using \ourmethod's performance on PASCAL MTL}
  \label{fig:ab5}
\end{figure}

\noindent\textbf{CPAB Parameterization.} 
\revision{The CPAB transformations are parameterized by the number of subintervals $\gN_\gP$ of the domain $\Omega$. From \Cref{fig:ab5}, we observe that a moderate-sized $\gN_\gP = 32$ provides strong and stable performance.}

\noindent\textbf{Pre-training Scale.} Models pre-trained on ImageNet-21k mostly outperform their ImageNet-1k counterparts (\Cref{tab:pretrainsize}), suggesting that models pre-trained on larger datasets learn novel representations and, by preserving them, make \ourmethod\ more effective.
\begin{table}[h]
    \centering
     \caption{MTL Performance using \ourmethod\ on PASCAL for varying pre-training dataset scale}
    \begin{tabular}{lcc}
       \toprule
       \textbf{Task} $\downarrow$ / \textbf{Dataset} $\rightarrow$ & \textbf{ImageNet-1k} & \textbf{ImageNet-21k} \\
       \midrule
       \textsc{SemSeg} & 70.09 & 69.06 \\
       \textsc{Human Parts} & 59.03 & 62.02\\
       \textsc{Saliency} & 64.55 & 65.00\\
       \textsc{Normals} & 17.47 & 17.10\\
       \bottomrule
    \end{tabular}
    \label{tab:pretrainsize}
\end{table}

\begin{figure}[t]
    \centering
    \includegraphics[width=\linewidth]{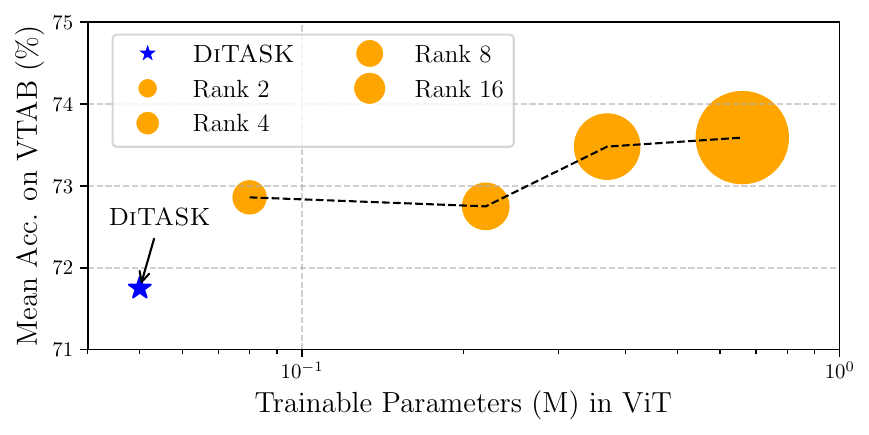}
    \caption{Pareto optimal curve on VTAB benchmark using \ourmethod\ and LoRA. \ourmethod\ achieves competitive performance using $\sim 10 \times$ fewer parameters.}
    \label{fig:VTAB}
\end{figure}

\revision{\noindent\textbf{VTAB Benchmark.} To understand the single-task generalization capabilities of our method, we compare \ourmethod\ and LoRA~\citep{hu2022lora} using the ViT~\citep{dosovitskiy2021an} backbone on the VTAB Benchmark~\citep{zhai2020largescalestudyrepresentationlearning}. Shown in \Cref{fig:VTAB}, \ourmethod\ achieves competitive performance with an order of magnitude fewer learnable parameters, presenting itself as a strong fine-tuning method for vision transformers.}

\revision{\noindent\textbf{Additional Experiments. }}
\revision{To understand the robustness of our design for different backbones, we evaluate \ourmethod\ using the Pyramid Vision Transformer (PVT)~\citep{wang2021pyramid} backbone against various parameter budgets of MTLoRA and LoRA. From \Cref{tab:diff_vit}, we conclude that our \ourmethod achieves 2.5$\times$ improvement in multi-task performance over best-performing baseline using 4.4$\times$ fewer trainable backbone parameters.}
\begin{table}[t]
  \centering
  \scriptsize
   \setlength{\tabcolsep}{2.5pt}
  \caption{
  MTL Performance of selected baselines vs. \ourmethod\ using Pyramid Vision Transformer (PVT) and Swin-Tiny backbones with different parameter budgets.
  }
  \vspace{-5pt}
  \begin{tabular}{l  c  c}
    \toprule
    \multirow{2}{*}{\textbf{Method}} & 
      \multirow{2}{*}{$\Delta m (\%) $} & \textbf{Trainable Backbone} \\
       & &
      \textbf{Parameters} (M)    \\
    \midrule
   PVT +  LoRA ($r=4$)  & -1.35 & 2.41 \\
   Swin-Tiny + LoRA ($r=4$) & -2.17 & 0.93\\
   Swin-Tiny + LoRA ($r=8$) &  +4.93 & 1.31 $(\times 4)$ \\
   \midrule 
    PVT + MTLoRA ($r = 64$) & +1.2 & 8.69 \\    
     Swin-Tiny + MTLoRA ($r = 16 $) & +1.35 & 3.01\\
     Swin-Tiny + MTLoRA ($r = 32 $) & +2.16 & 4.14\\
     Swin-Tiny + MTLoRA ($r = 64 $) & +2.55 & 6.40\\
     \midrule 
       PVT + \ourmethod &  \textbf{+3.01} & 1.96 \\
        Swin-Tiny + \ourmethod &  \textbf{+3.22} & 1.61\\
        (Single Task) Swin-Tiny + \ourmethod & \textbf{+5.33} & 1.61 $(\times 4)$\\
    \bottomrule
  \end{tabular}
  \label{tab:diff_vit}
  \vspace{-12pt}
\end{table}\\

  
  
\begin{figure}[!htbp]
\centering	\newcommand{\imwidth}{0.12\textwidth}
	\setlength{\tabcolsep}{1pt}
	\begin{tabular}{ccccccc}
	\toprule
\multicolumn{4}{c}{Semantic Segmentation} \\ 
\midrule
\shortstack{Input Image} &
\shortstack{MTLoRA} & 
\shortstack{{\ourmethod}} & 
\shortstack{{Ground Truth}} \\
\midrule 
\includegraphics[width=\imwidth]{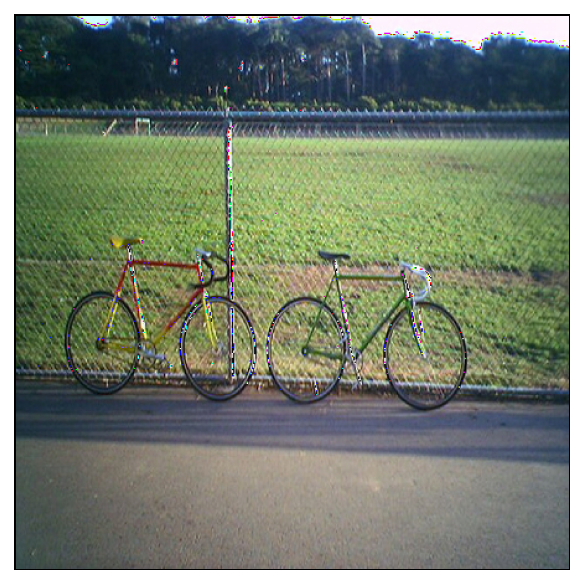} & \includegraphics[width=\imwidth]{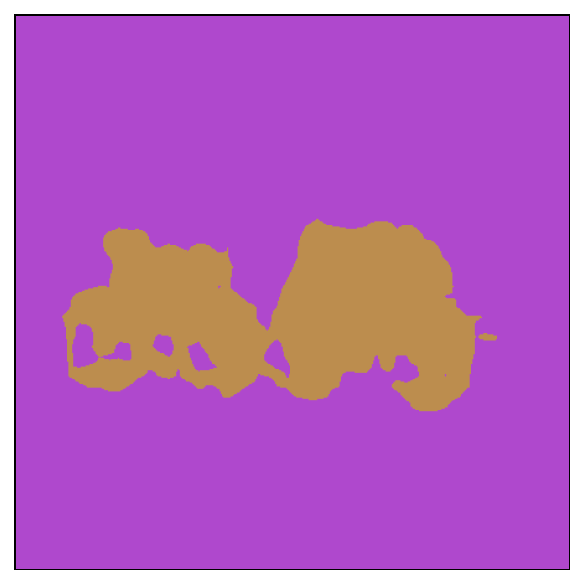} &
\includegraphics[width=\imwidth]{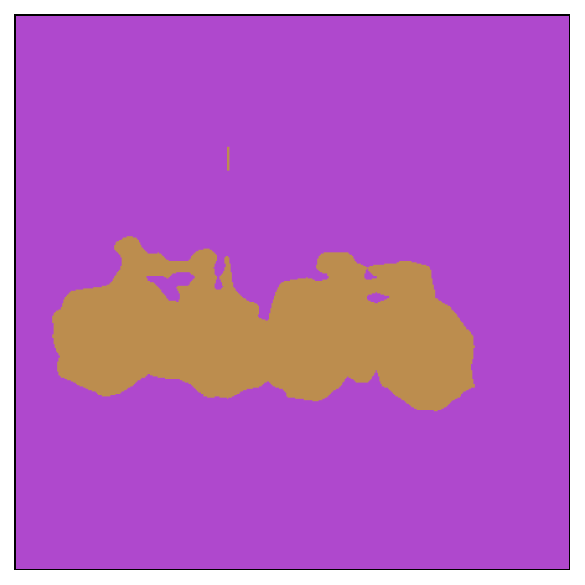}
& 
\includegraphics[width=\imwidth]{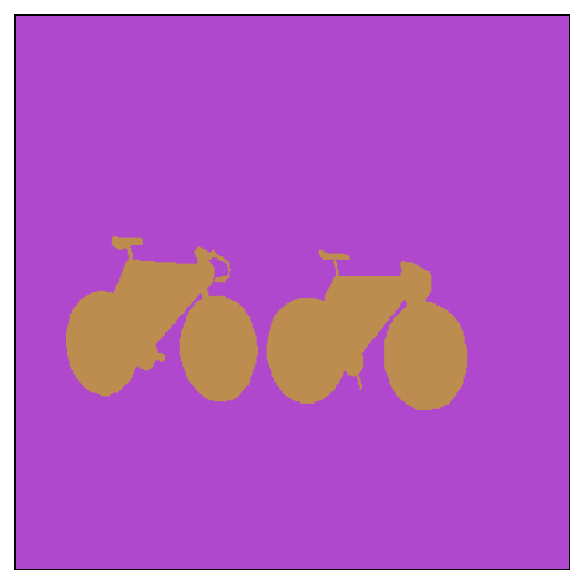}\\

\includegraphics[width=\imwidth]{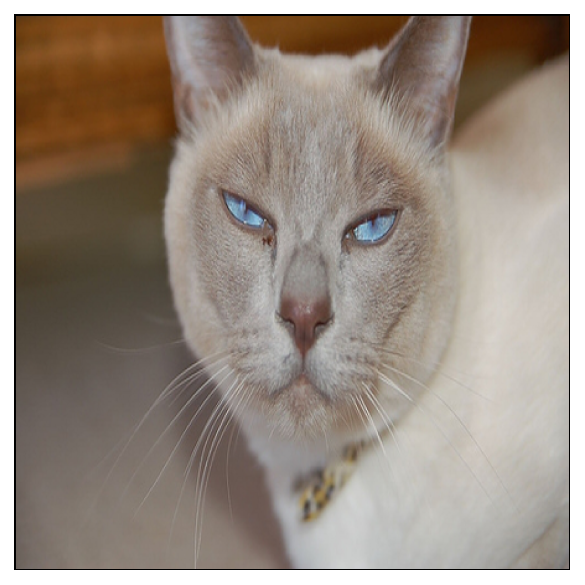} & \includegraphics[width=\imwidth]{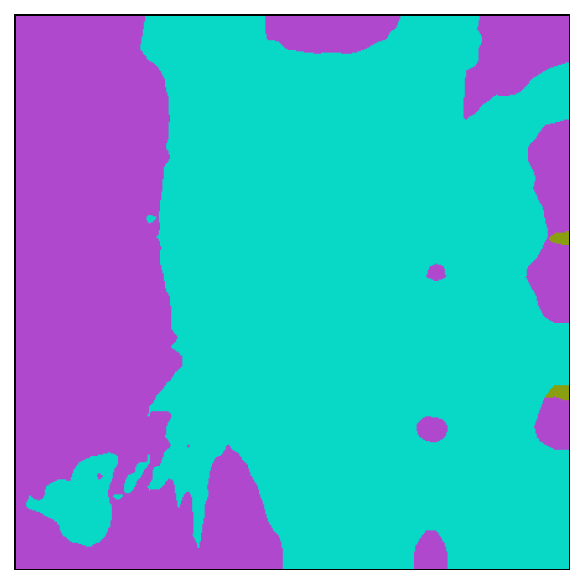} &
\includegraphics[width=\imwidth]{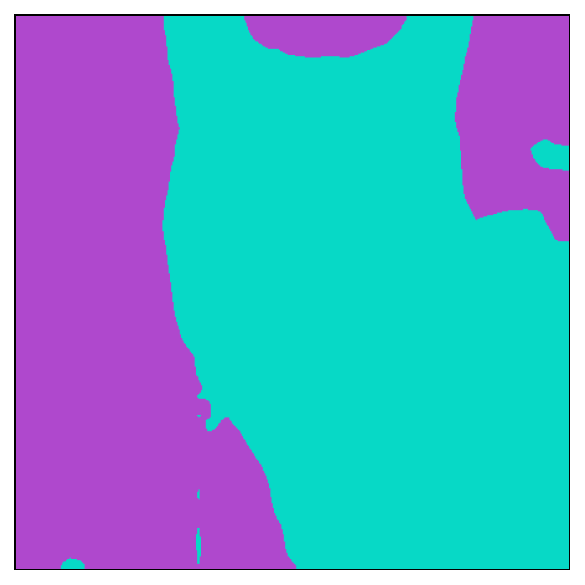}
& 
\includegraphics[width=\imwidth]{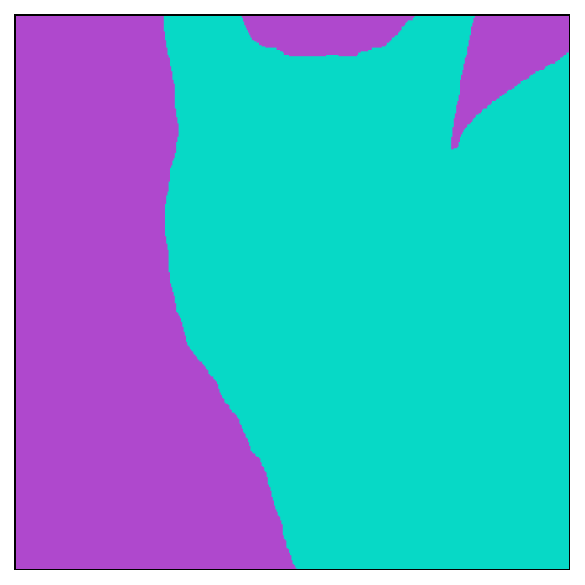}\\

\includegraphics[width=\imwidth]{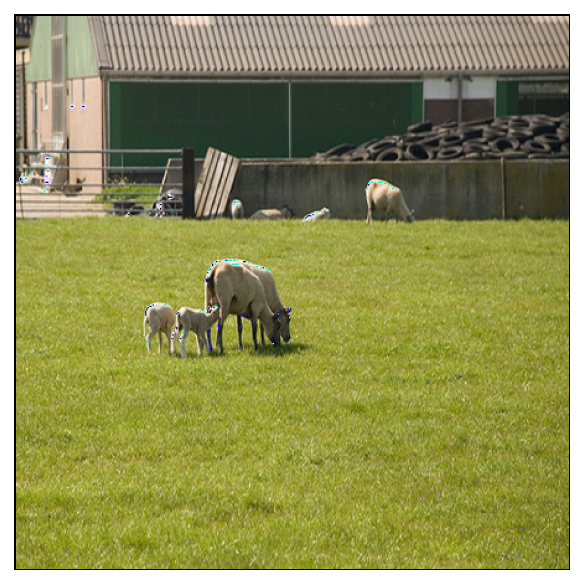} & \includegraphics[width=\imwidth]{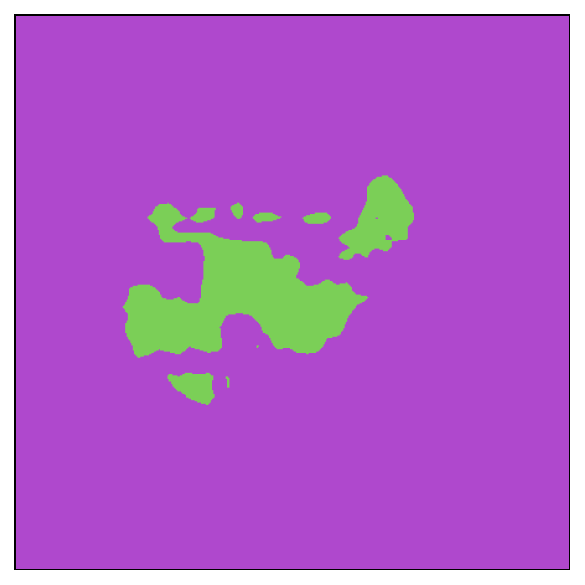} &
\includegraphics[width=\imwidth]{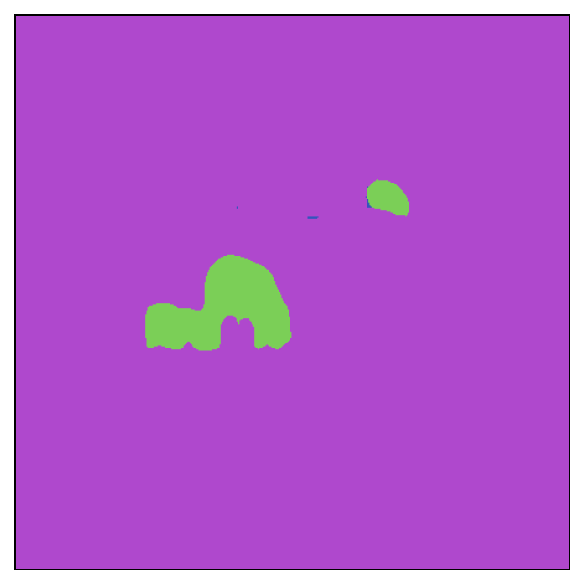}
& 
\includegraphics[width=\imwidth]{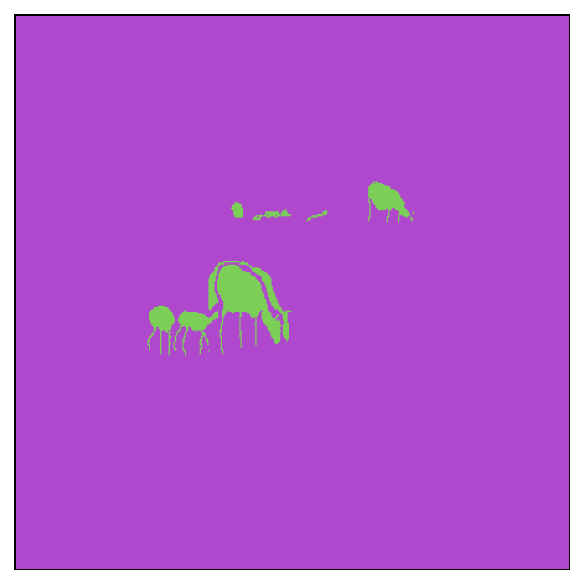}\\

\includegraphics[width=\imwidth]{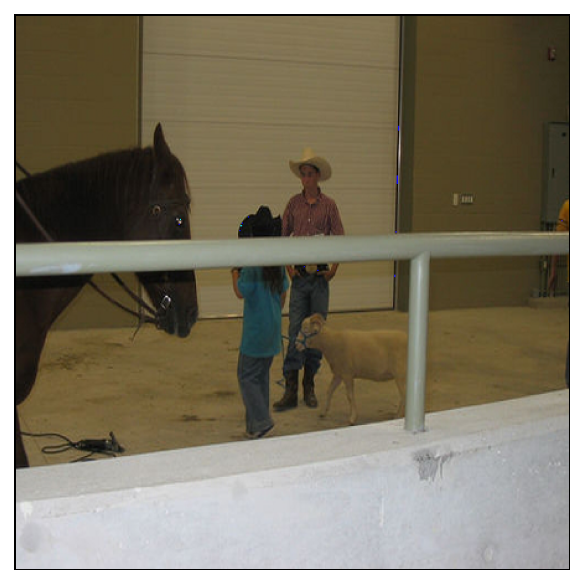} & \includegraphics[width=\imwidth]{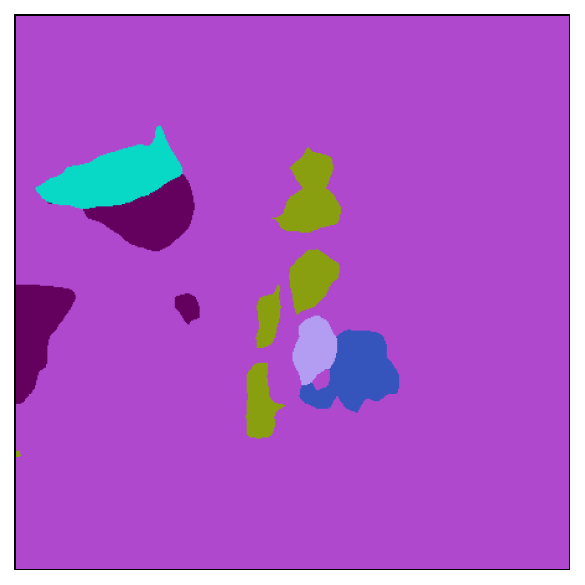} &
\includegraphics[width=\imwidth]{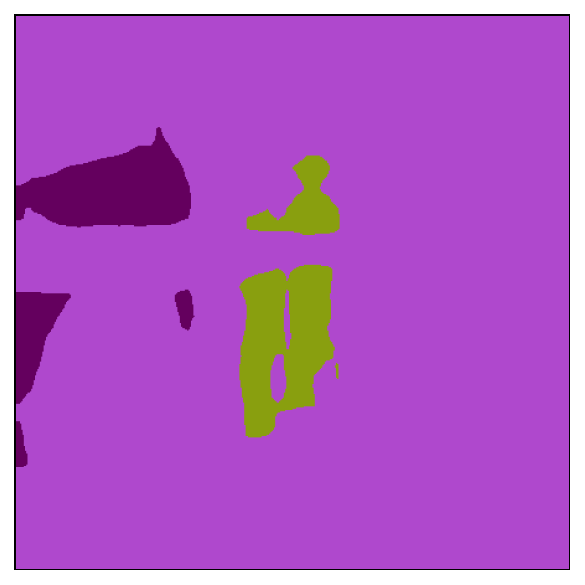}
& 
\includegraphics[width=\imwidth]{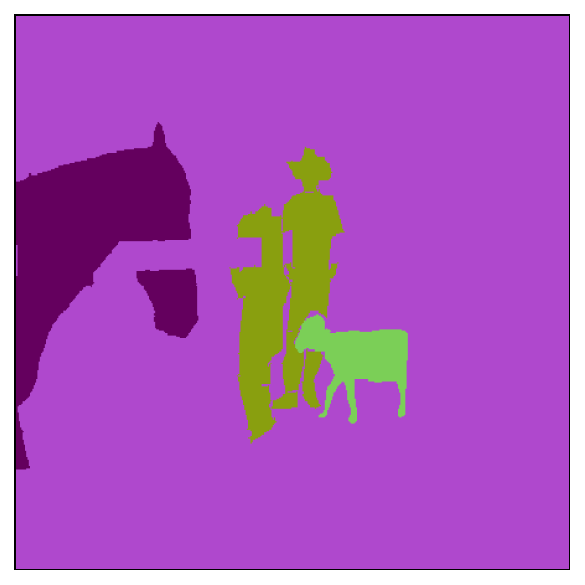}\\

\includegraphics[width=\imwidth]{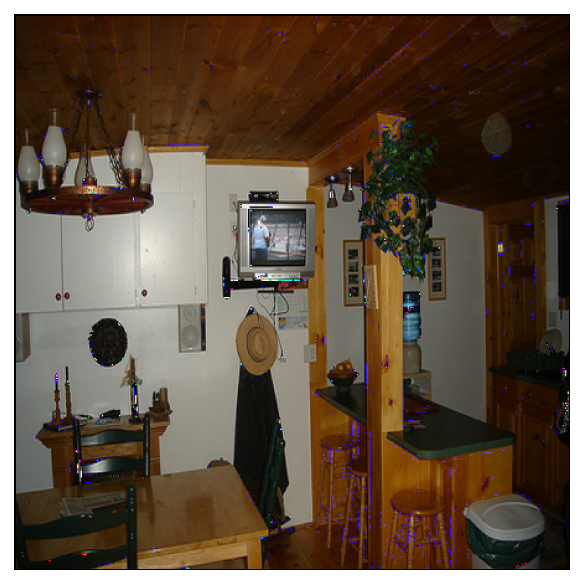} & \includegraphics[width=\imwidth]{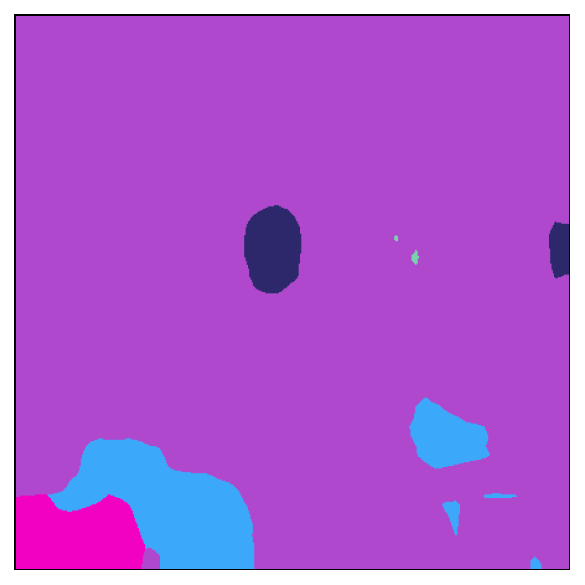} &
\includegraphics[width=\imwidth]{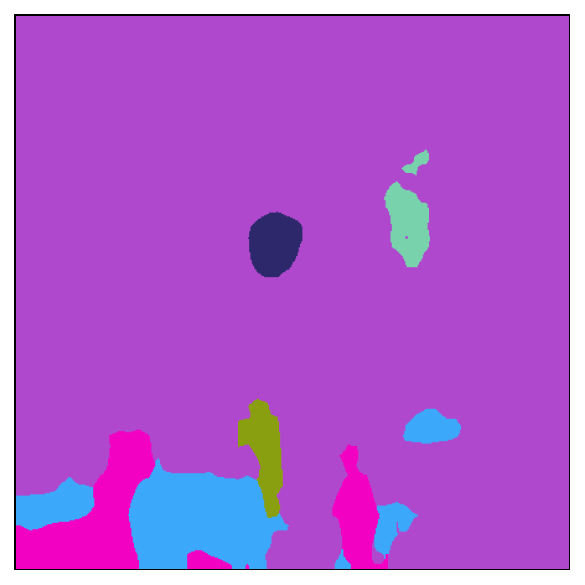}
& 
\includegraphics[width=\imwidth]{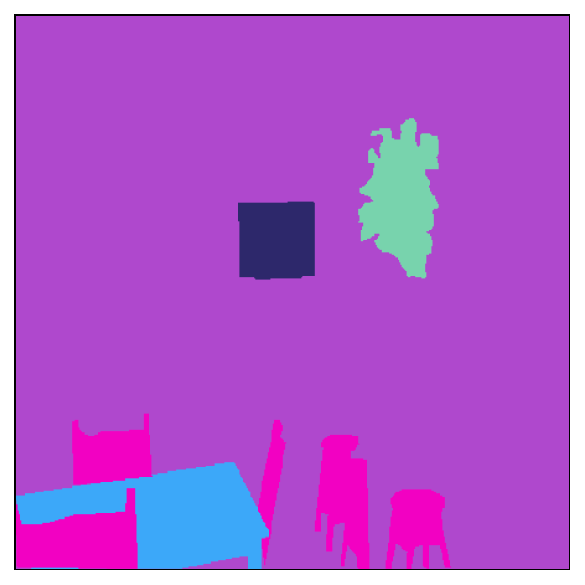}\\

\includegraphics[width=\imwidth]{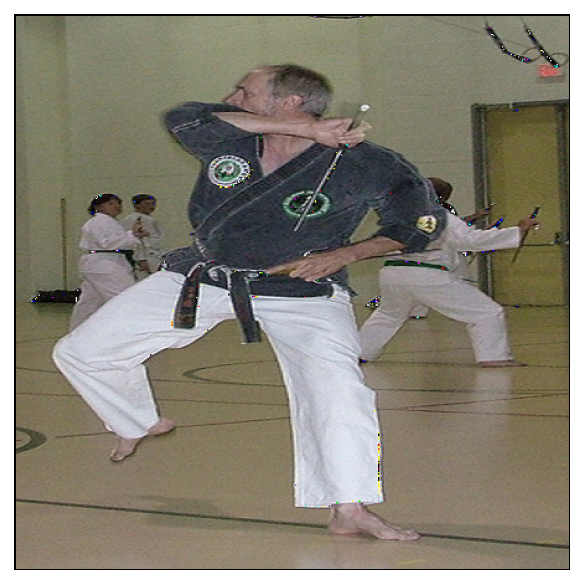} & \includegraphics[width=\imwidth]{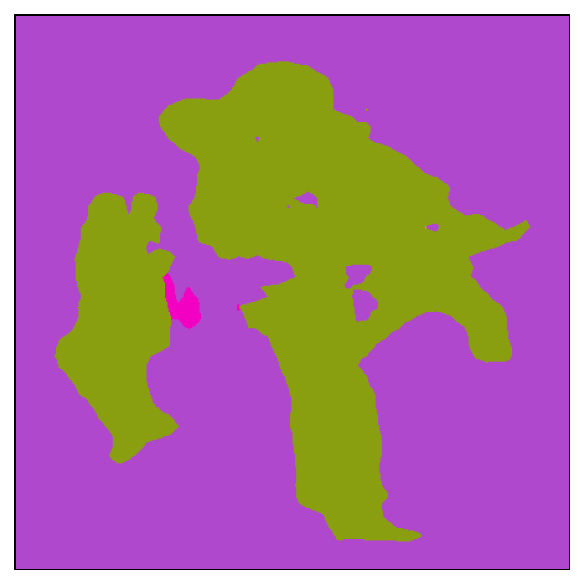} &
\includegraphics[width=\imwidth]{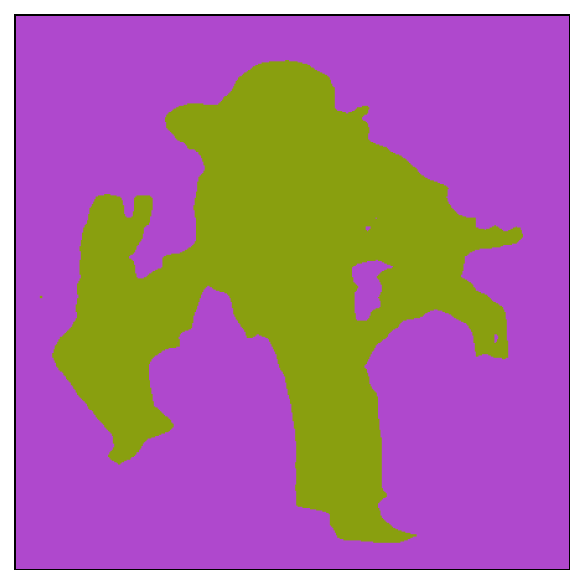}
& 
\includegraphics[width=\imwidth]{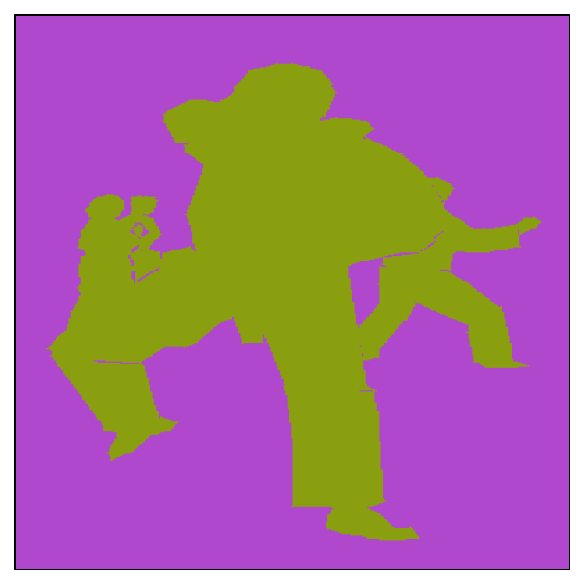}\\

\includegraphics[width=\imwidth]{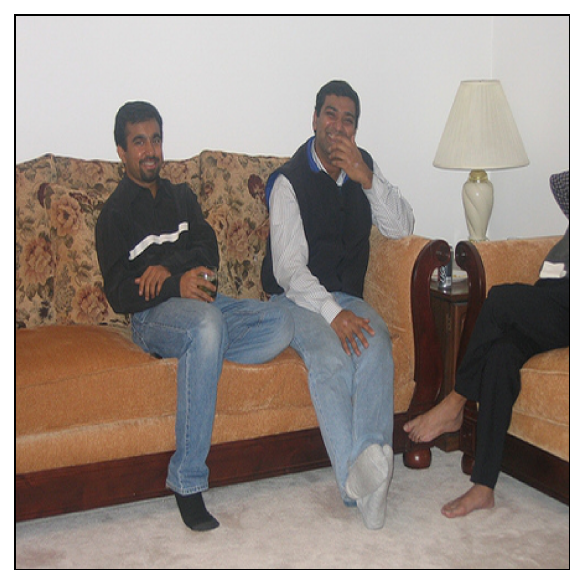} & \includegraphics[width=\imwidth]{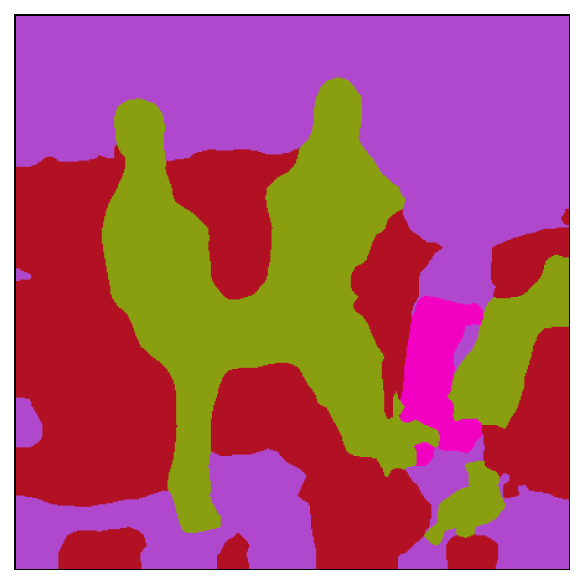} &
\includegraphics[width=\imwidth]{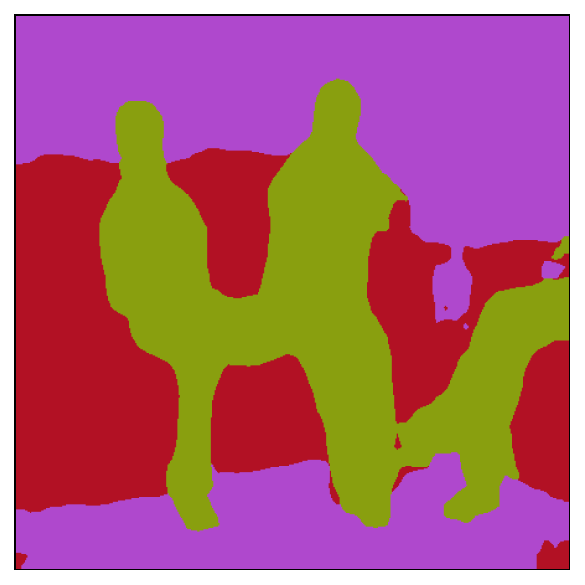}
& 
\includegraphics[width=\imwidth]{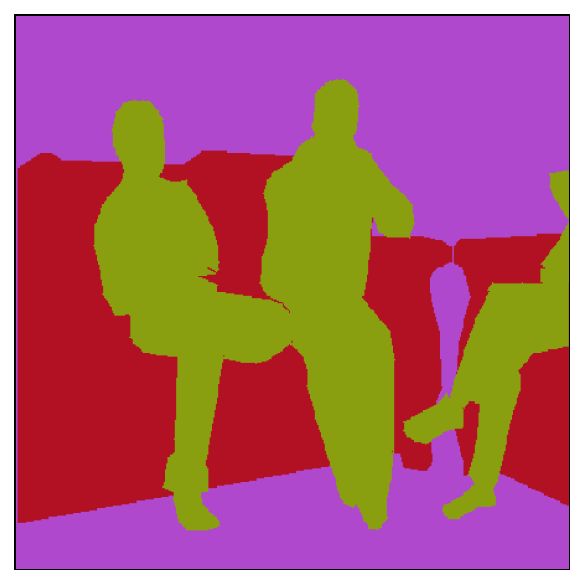}\\

\includegraphics[width=\imwidth]{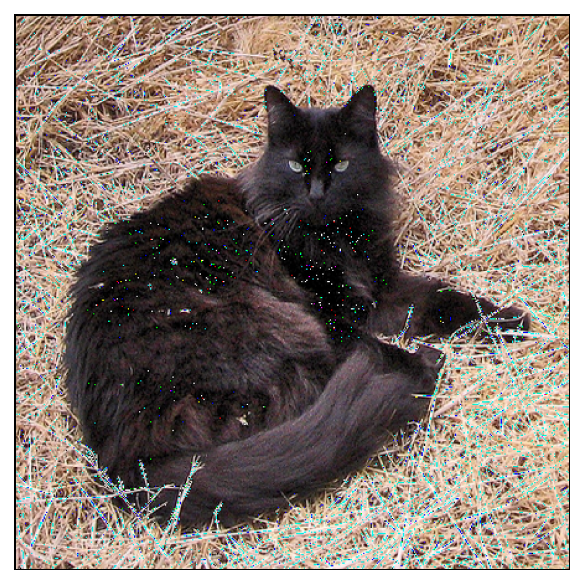} & \includegraphics[width=\imwidth]{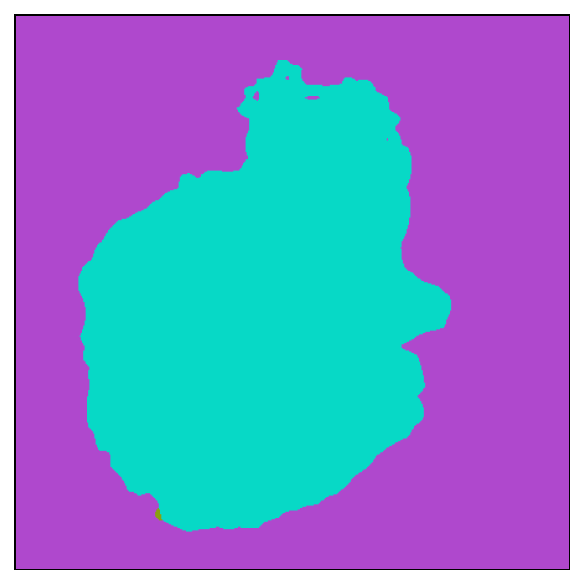} &
\includegraphics[width=\imwidth]{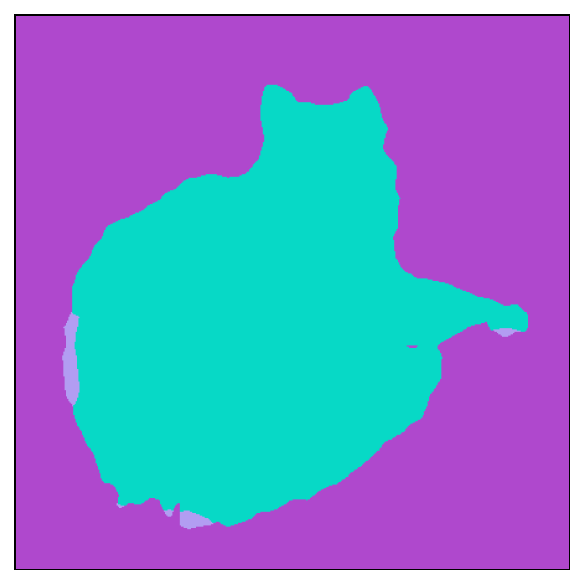}
& 
\includegraphics[width=\imwidth]{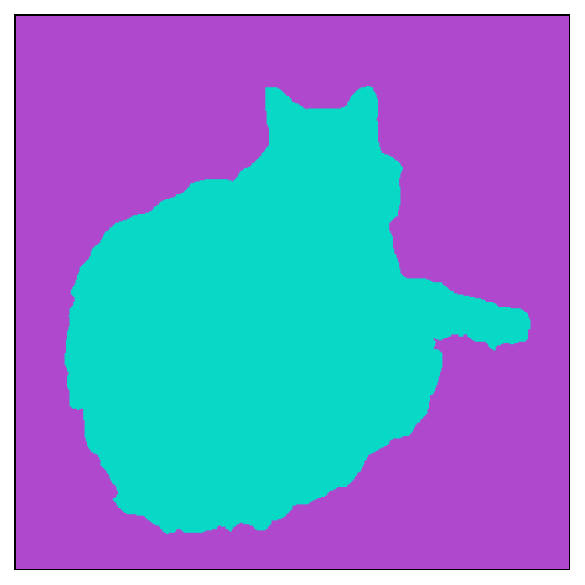}\\

\bottomrule
\end{tabular}
\captionof{figure}{Qualitative comparison of semantic segmentation on representative samples from the PASCAL MTL dataset with MTLoRA and our \ourmethod}
\label{fig:viz3}
\end{figure}

\section{Qualitative Comparison}


\Cref{fig:viz3} shows semantic segmentation results on the PASCAL MTL~\citep{pascal} dataset, comparing MTLoRA and \ourmethod\ against the ground truth. \ourmethod\ consistently produces sharper and more accurate segmentations. In the first row, it captures the structure and boundaries of bicycles more precisely, whereas MTLoRA over-smooths the outputs, failing to recover fine details. Similarly, in the third row, \ourmethod\ segments smaller objects, such as zebras, with greater detail, avoiding omissions observed in MTLoRA. For complex indoor scenes, such as rows five and six, \ourmethod\ distinguishes multiple objects effectively and maintains segmentation coherence, whereas MTLoRA generates fragmented outputs. These results highlight \ourmethod’s ability to adapt pre-trained weights through diffeomorphic transformations, enabling accurate segmentation across diverse object categories.


\begin{figure}[!t]
\centering
	\newcommand{\imwidth}{0.12\textwidth}
	\setlength{\tabcolsep}{1pt}
	\begin{tabular}{ccccccc}
	\toprule
\multicolumn{4}{c}{Depth Estimation} \\ 
\midrule
\shortstack{Input Image} &
\shortstack{MTLoRA} & 
\shortstack{{\ourmethod}} & 
\shortstack{{Ground Truth}} \\
\midrule 
\includegraphics[width=\imwidth]{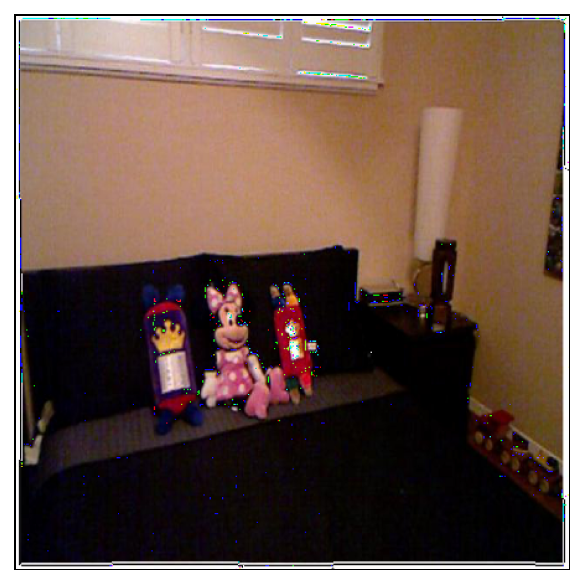} & \includegraphics[width=\imwidth]{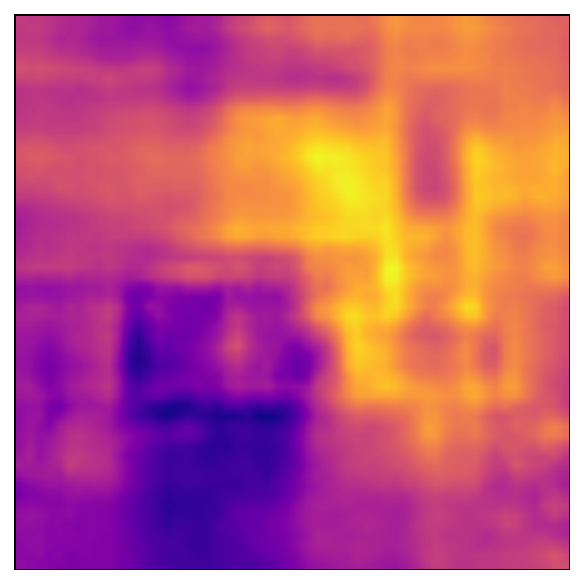}&
\includegraphics[width=\imwidth]{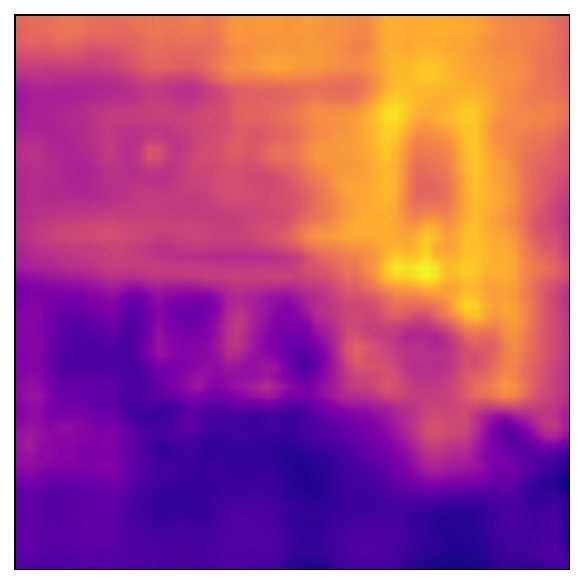}
& 
\includegraphics[width=\imwidth]{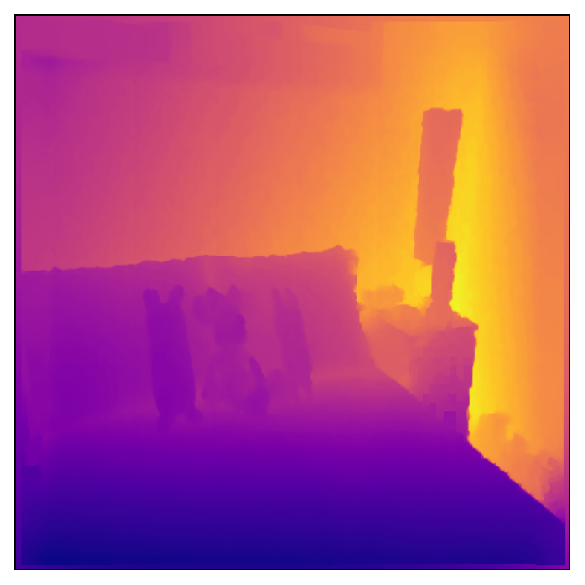}\\
\includegraphics[width=\imwidth]{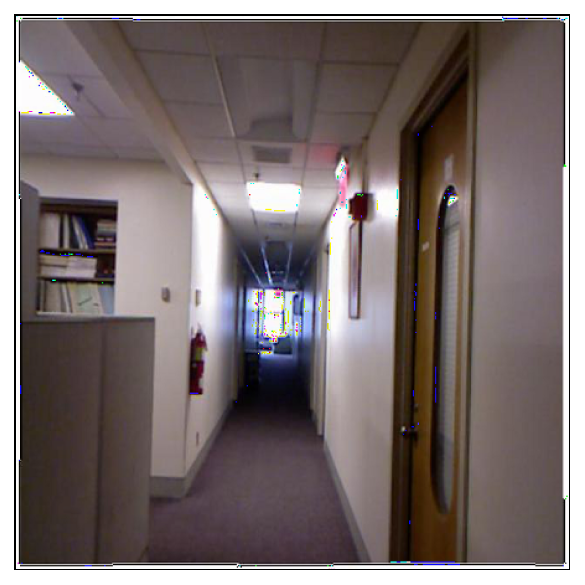} & \includegraphics[width=\imwidth]{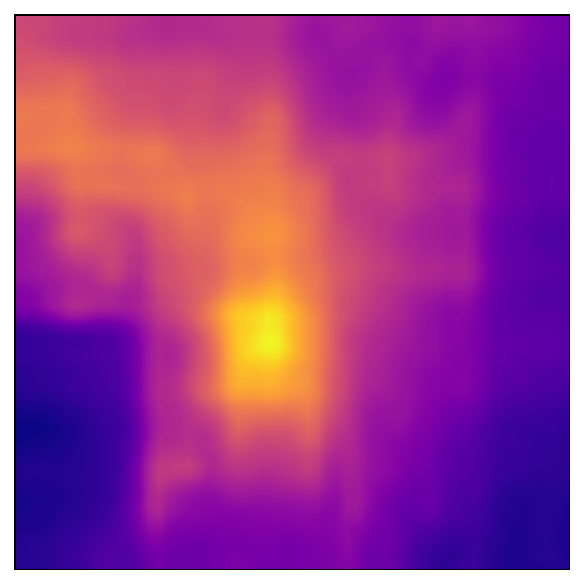}&
\includegraphics[width=\imwidth]{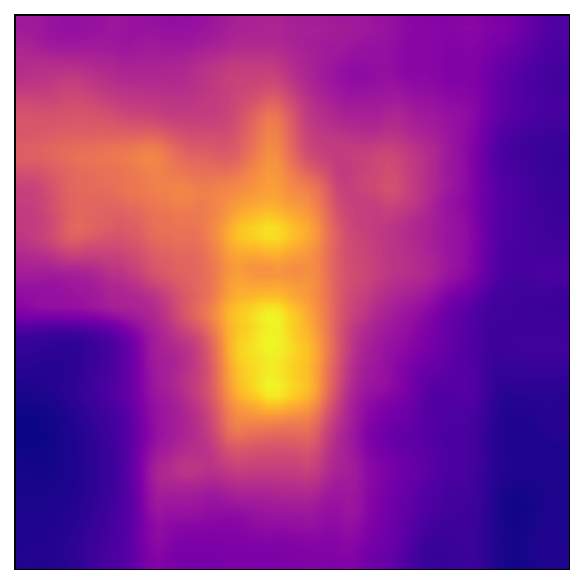}
& 
\includegraphics[width=\imwidth]{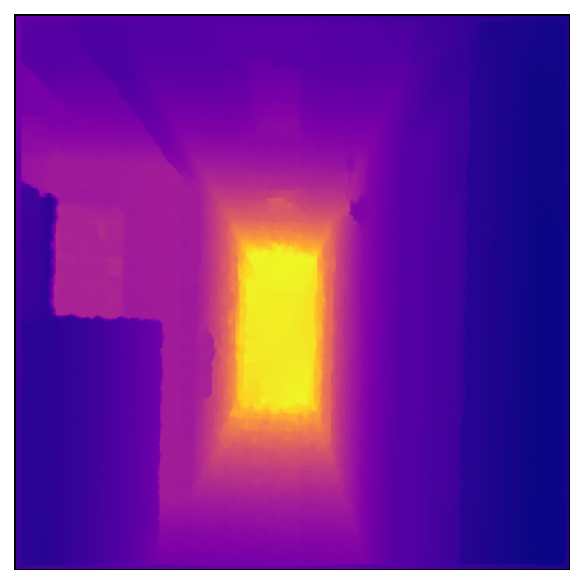}\\
\includegraphics[width=\imwidth]{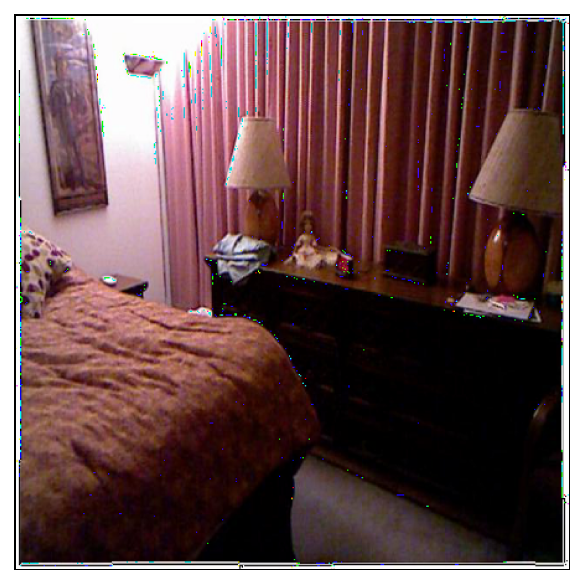} & \includegraphics[width=\imwidth]{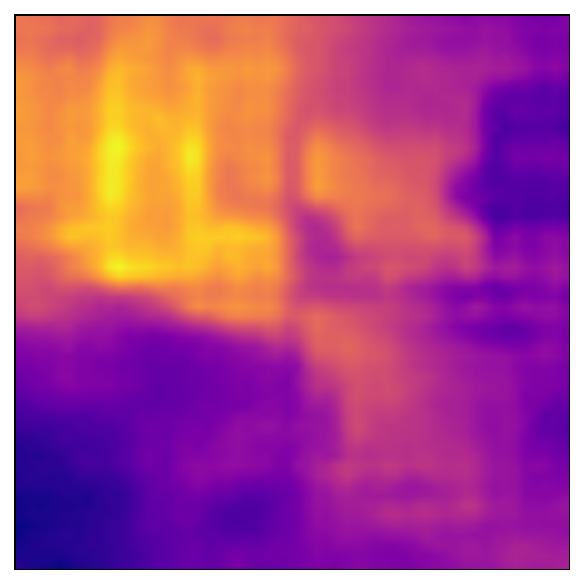}&
\includegraphics[width=\imwidth]{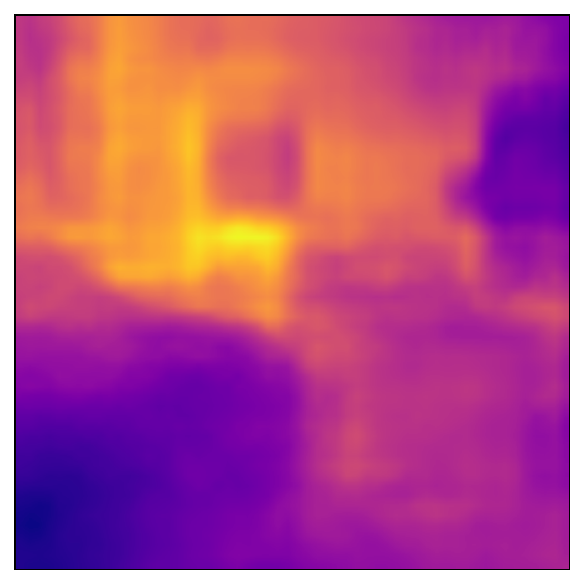}
& 
\includegraphics[width=\imwidth]{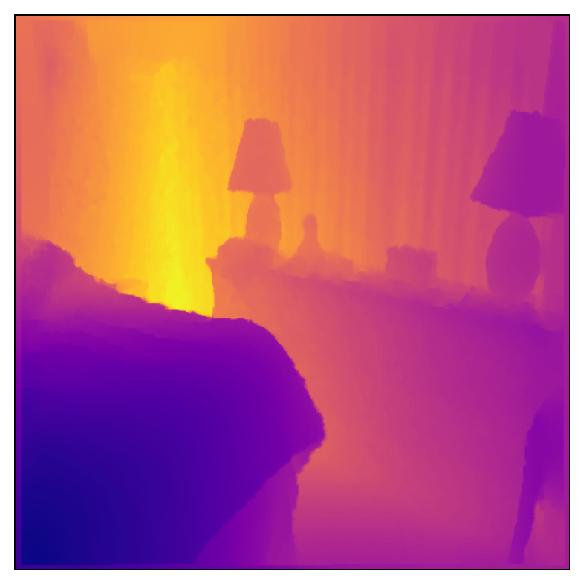}\\
\includegraphics[width=\imwidth]{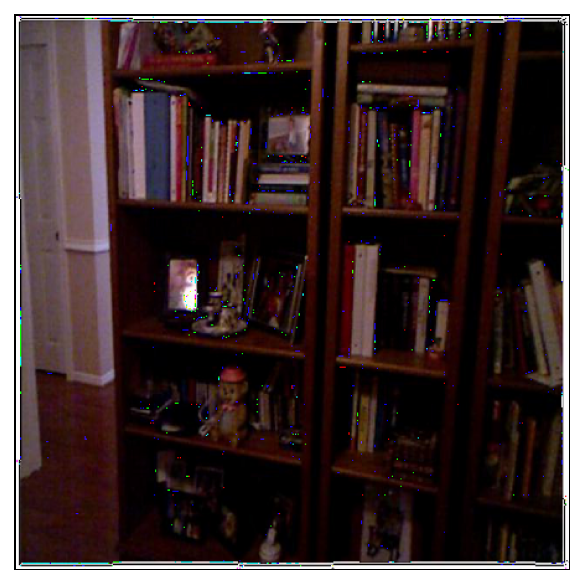} & \includegraphics[width=\imwidth]{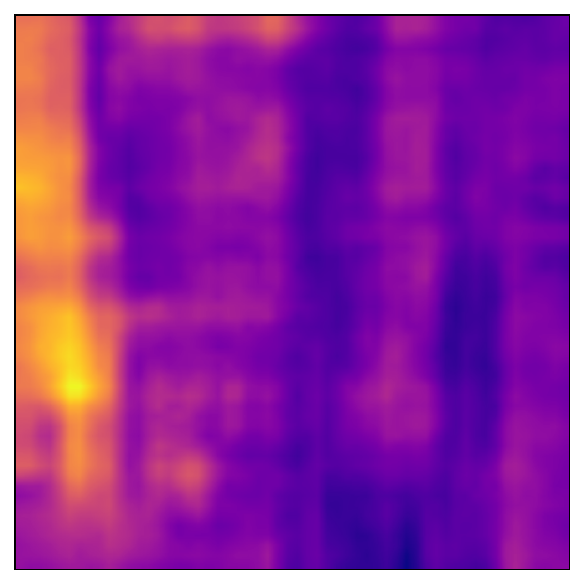}&
\includegraphics[width=\imwidth]{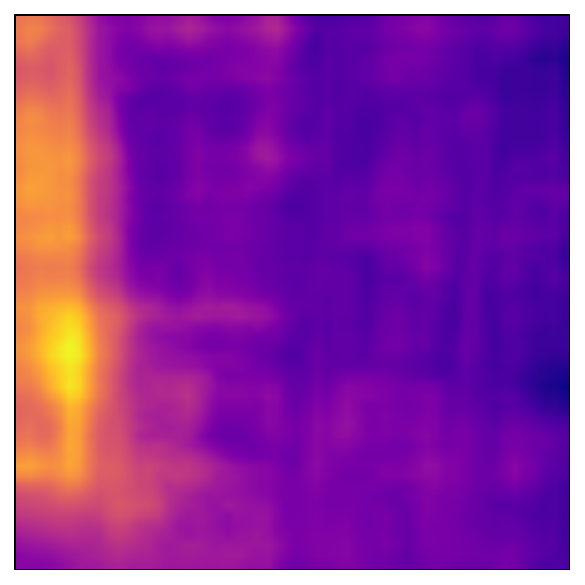}
& 
\includegraphics[width=\imwidth]{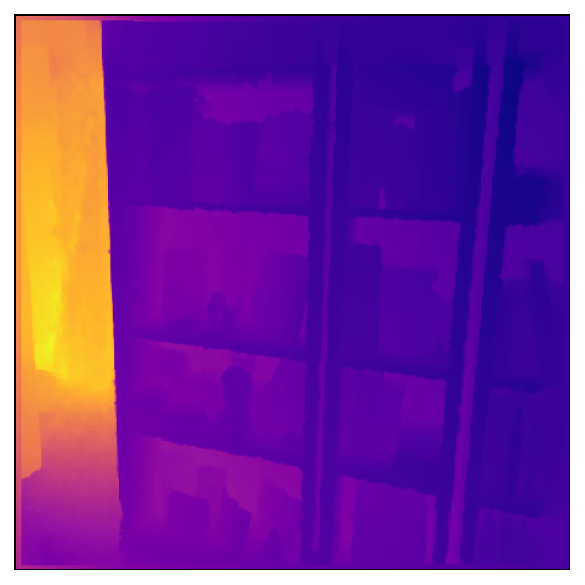}\\
\includegraphics[width=\imwidth]{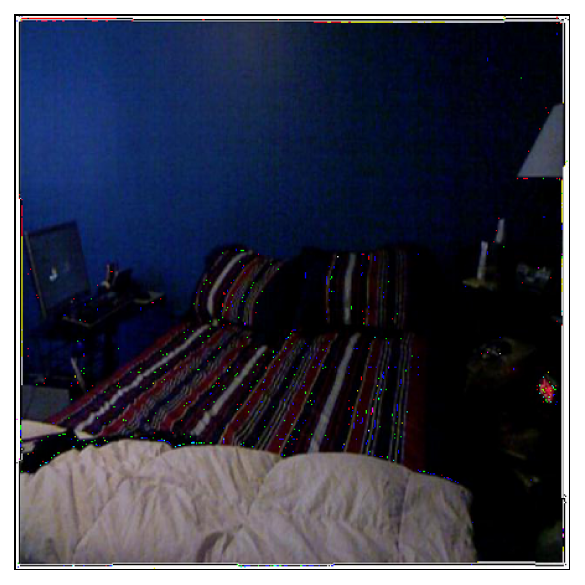} & \includegraphics[width=\imwidth]{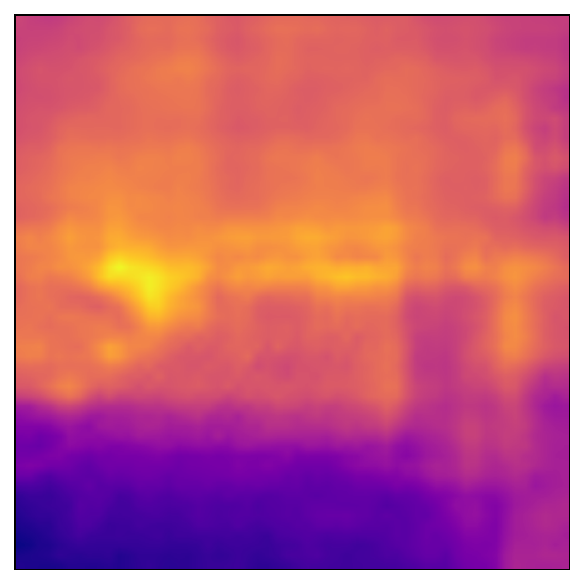}&
\includegraphics[width=\imwidth]{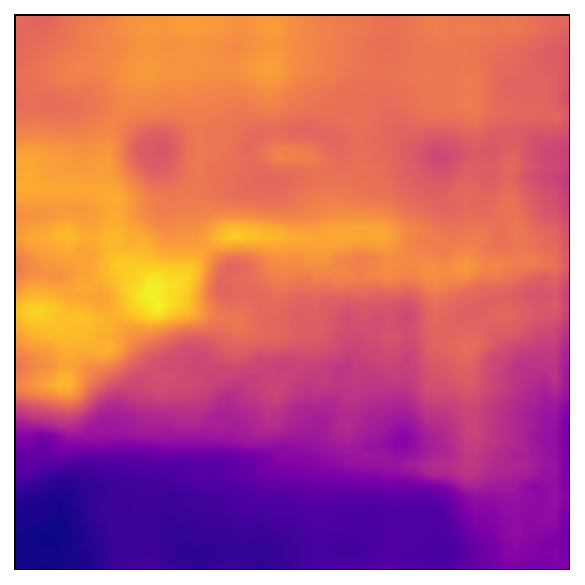}
& 
\includegraphics[width=\imwidth]{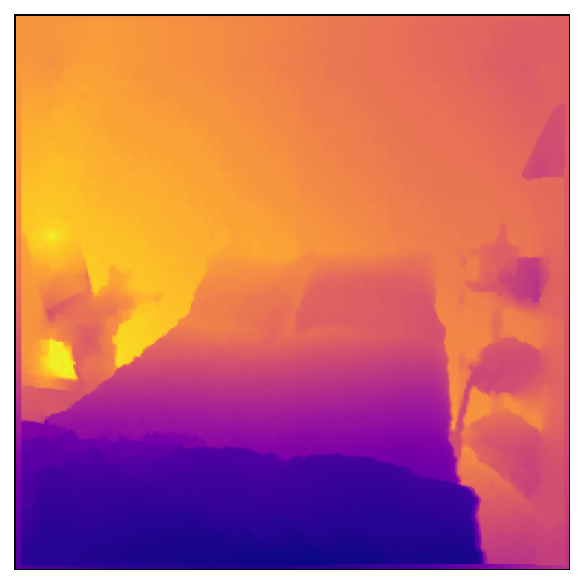}\\
\includegraphics[width=\imwidth]{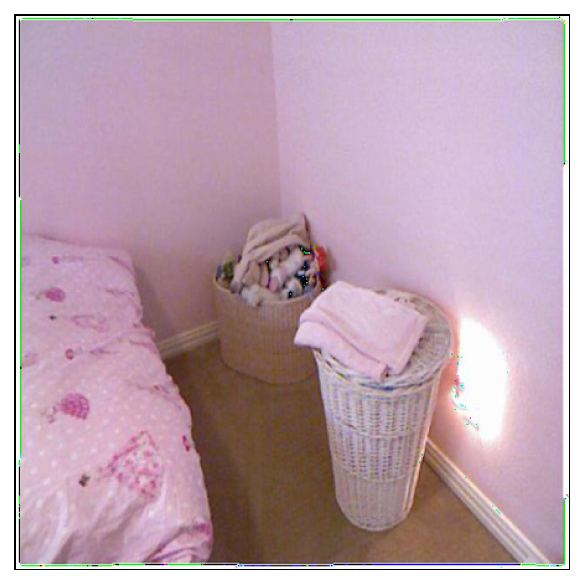} & \includegraphics[width=\imwidth]{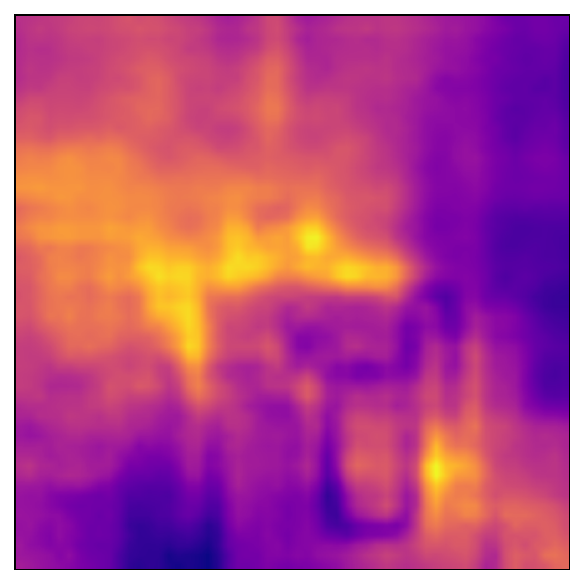}&
\includegraphics[width=\imwidth]{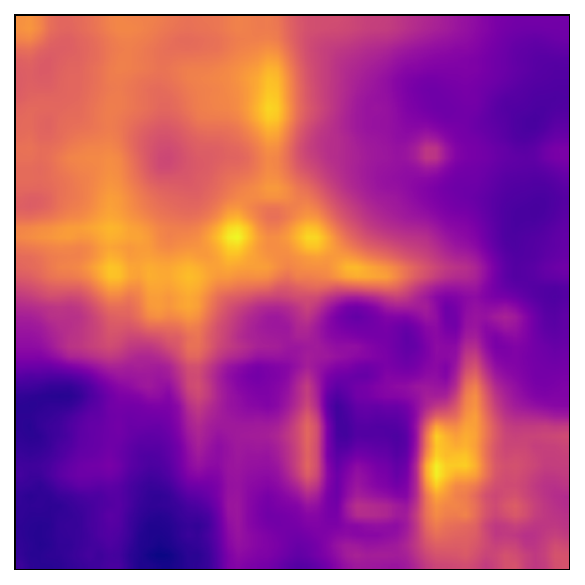}
& 
\includegraphics[width=\imwidth]{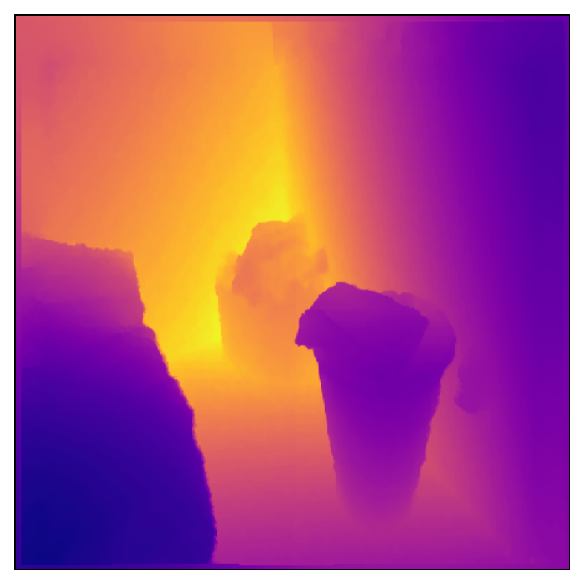}\\
\includegraphics[width=\imwidth]{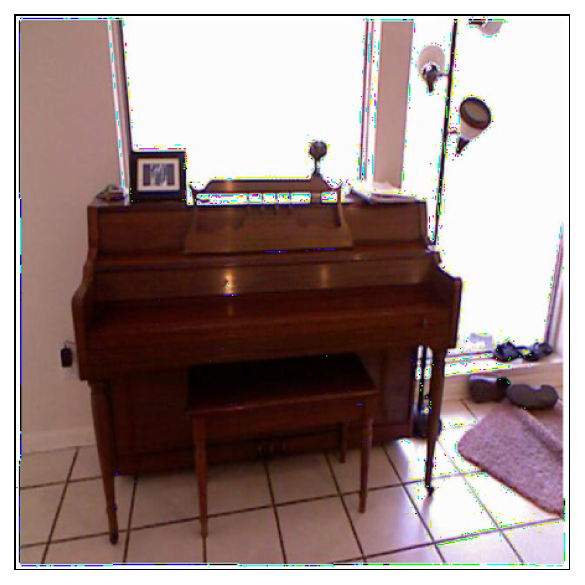} & \includegraphics[width=\imwidth]{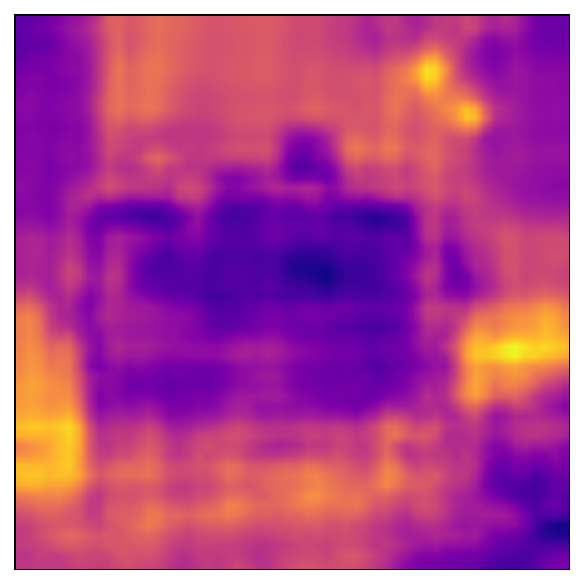}&
\includegraphics[width=\imwidth]{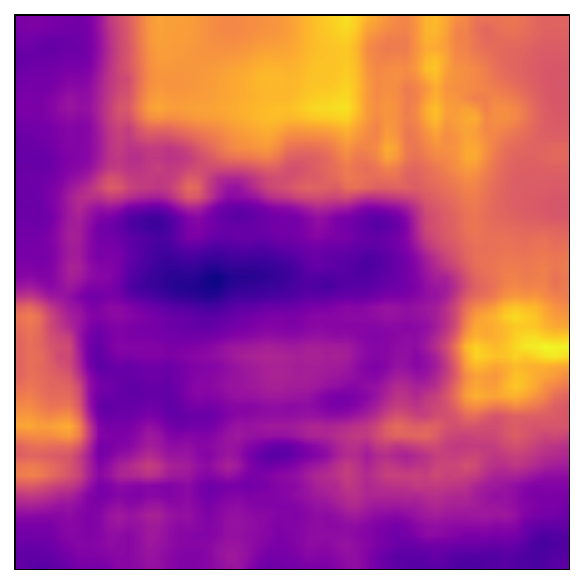}
& 
\includegraphics[width=\imwidth]{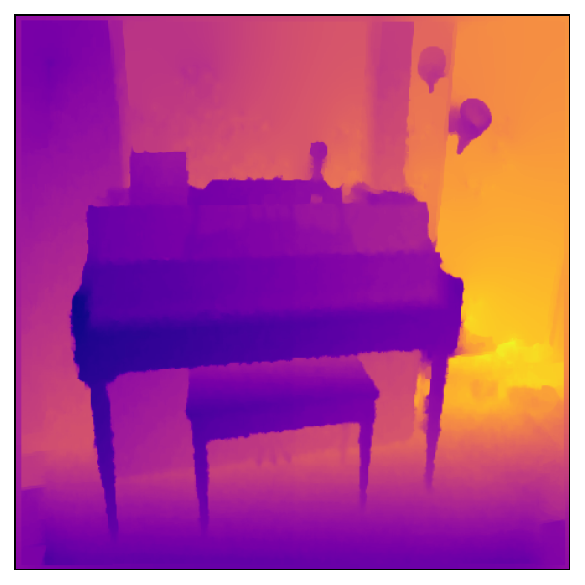}\\

\bottomrule
\end{tabular}
\captionof{figure}{Qualitative comparison of depth estimation on representative samples from the NYUD MTL dataset with MTLoRA and our \ourmethod}
\label{fig:viz2}
\end{figure}

\Cref{fig:viz2} demonstrates depth estimation results on the NYUD MTL~\citep{silberman2012indoor} dataset, comparing MTLoRA and \ourmethod. \ourmethod\ captures fine-grained depth variations and preserves object boundaries more effectively than MTLoRA. In the second row, \ourmethod\ separates the hallway’s foreground and background accurately, closely matching the ground truth, while MTLoRA produces oversimplified and blurred outputs. In the fourth row, \ourmethod\ preserves depth discontinuities and object structures, where MTLoRA fails to capture these transitions. Even in challenging scenes, such as the last row, \ourmethod\ achieves detailed and consistent depth predictions, outperforming MTLoRA. These results validate the effectiveness of \ourmethod’s singular value transformations in producing precise, task-specific depth estimates.

\revision{\section{Additional Related Work}}

\noindent\textbf{\revision{Hard Parameter Sharing and Task Dynamics in Multi-Task Learning. }
} \revision{Hard parameter sharing is a widely used approach in multi-task learning (MTL), where most layers of a neural network are shared among tasks, while task-specific layers are restricted to the output heads. This technique, introduced in \citep{caruana1997multitask}, is computationally efficient but presents challenges due to task interference, where conflicting task gradients degrade performance. Approaches like PCGrad~\citep{yu2020gradient} mitigate this by enforcing gradient orthogonality, but they do not always address the full extent of task competition for limited shared parameters. Despite these challenges, task synergies can be harnessed through careful parameter modulation, allowing shared features to benefit related tasks. In this context, methods like our \ourmethod\ enhance positive transfer by preserving crucial pre-trained feature structures, enabling both task-agnostic and task-specific adaptations through diffeomorphic transformations. }

\noindent\textbf{\revision{Paradigms in Multi-Task Learning: Model Design and Optimization Strategies. }}
\revision{Research on MTL can be categorized into two main paradigms: optimization-driven and model design-based strategies. Optimization approaches focus on balancing task-specific loss functions or modifying task gradients to reduce interference, as seen in works like PCGrad~\citep{yu2020gradient}. In contrast, model design-based methods, such as adapters~\citep{he2021towards} and LoRA~\citep{hu2022lora}, introduce parameter-efficient layers that balance shared and task-specific features. However, these methods often restrict updates to low-rank subspaces, limiting adaptability. MTLoRA~\citep{agiza2024mtlora} extends LoRA by incorporating task-specific subspaces, yet still faces trade-offs between task isolation and synergy. Our \ourmethod\ addresses these limitations by preserving full-rank features and enabling dynamic, parameter-efficient adaptations, achieving superior performance on MTL benchmarks.}

\end{document}